\documentclass[conference]{IEEEtran}
\usepackage{times}

\usepackage[numbers, sort]{natbib}
\usepackage{multicol}
\usepackage[bookmarks=true]{hyperref}

\usepackage[nolist]{acronym}
\usepackage[cmex10]{amsmath}
\usepackage{amssymb}
\usepackage{bm}
\usepackage{booktabs}
\usepackage{color}
\usepackage{xspace}
\usepackage{amsmath}
\usepackage{algorithm}
\usepackage{algpseudocode}
\usepackage{algorithmicx}

\usepackage{float}
\usepackage{graphicx}
\usepackage{hyperref}
\usepackage{import}
\usepackage{mathtools}
\usepackage{multirow}
\usepackage{outline}
\usepackage{paralist}
\usepackage{siunitx}
\usepackage{stmaryrd}
\usepackage{subfig}
\usepackage{systeme}
\usepackage{url}
\usepackage{tikz}
\usepackage{booktabs}
\usepackage{cancel}
\usepackage{etoolbox}
\usepackage{soul}

\newtheorem{remark}{Remark}

\begin{document}

\newcommand{\method}{DiffuseLoco\xspace}

\newcommand{\dt}{TF\xspace}

\newcommand{\ddim}{DDIM-100/10\xspace}

\newcommand{\ddimn}{DDIM-10/5\xspace}

\newcommand{\fixme}[1]{{\color{red} {#1}}}

\newcommand{\xh}[1]{\textcolor{cyan}{#1}}

\newcommand{\yc}[1]{\textcolor{blue}{#1}}

\newcommand{\changed}[1]{\textcolor{black}{#1}}

\newcommand{\rw}[1]{\textcolor{lime}{#1}}

\makeatletter
\apptocmd{\@maketitle}{~~~~~~~~~~~~~~~~~~~~~~~~~~~~~~~~~~~~~~~~~~~~~~~~~~~~~~~~~~~~~~~~~~~~~~~~~~\insertfig}{}{} % hack to make the caption aligned to left
\makeatother

\newcommand{\insertfig}{%
  \makebox[0pt]{\includegraphics[width=\linewidth]{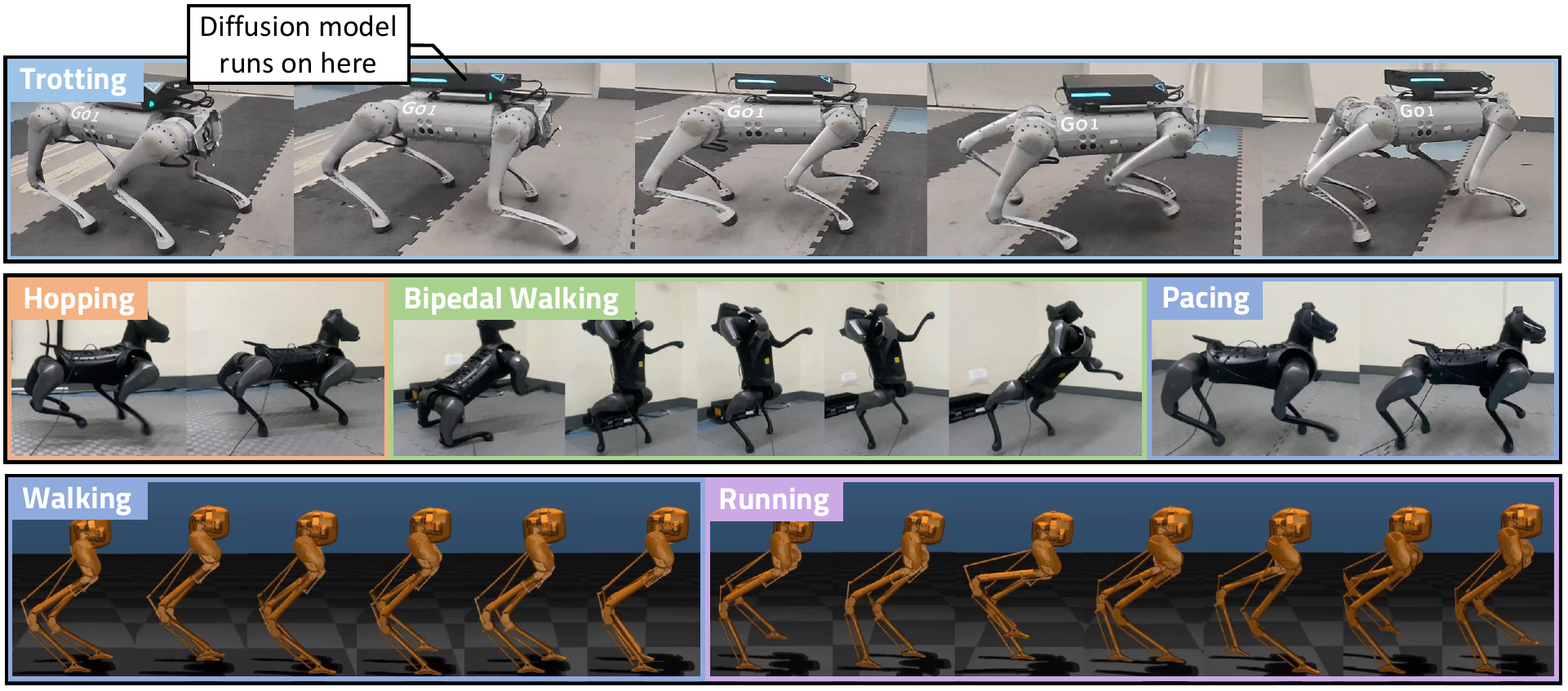}\label{fig:cover}} \\
  \small{Fig. 1. Snapshots of (top) a quadrupedal robot trotting with our diffusion-based real-time locomotion control policy via onboard computing; \changed{(middle) our multi-skill offline-learned policy performing a sequence of challenging skills, including hopping, bipedal locomotion, and pacing with smooth skill transitioning;  (bottom) a bipedal robot controlled by our multi-skill policy transitioning from walking to running stably.} We present \method, a scalable framework that leverages diffusion models to learn legged locomotion control exclusively from offline datasets. \method learns a state-of-the-art policy that is able to perform a diverse set of agile locomotion skills with a single policy, exhibiting robustness in the real world, and is versatile to various sources of offline data. \changed{We encourage the viewers to watch supplementary videos on these runs.}}
}

\title{DiffuseLoco: Real-Time Legged Locomotion Control with Diffusion from Offline Datasets} 

% \title{DiffuseLoco: Learning Real-Time Legged Locomotion Control from Offline Datasets by Diffusion}
% \title{DiffuseLoco: Diffusing Real-Time Legged Locomotion Control from Offline Datasets}

% You will get a Paper-ID when submitting a pdf file to the conference system
% \author{Author Names Omitted for Anonymous Review. Paper-ID [421]}
% \author{
%     Xiaoyu Huang$^{*, 1}$, 
%     Yufeng Chi$^{*, 1}$, 
%     Ruofeng Wang$^{*, 1}$, 
%     Zhongyu Li$^{1}$, 
%     Xue Bin Peng$^{2}$, 
%     Sophia Shao$^{1}$, 
%     Borivoje Nikolic$^{1}$, \\
%     Koushil Sreenath$^{1}$, 
%     \thanks{*Equal contribution.}%
%     \thanks{$^{1}$UC Berkeley, CA, USA.}%
%     \thanks{$^{2}$Simon Fraser University, BC, Canada.}%
%     \thanks{Contact: x.h@berkeley.edu}
% }
\author{\authorblockN{Xiaoyu Huang$^{*, 1}$,
Yufeng Chi$^{*, 1}$,
Ruofeng Wang$^{*, 1}$, 
Zhongyu Li$^{1}$, 
Xue Bin Peng$^{2}$, \\
Sophia Shao$^{1}$, 
Borivoje Nikolic$^{1}$, 
Koushil Sreenath$^{1}$}
\authorblockA{$^{1}$ UC Berkeley $^{2}$ Simon Fraser University}
\authorblockA{*Equal contribution. Email: \href{mailto:x.h@berkeley.edu}{x.h@berkeley.edu} }
}
%\author{\authorblockN{Michael Shell}
%\authorblockA{School of Electrical and\\Computer Engineering\\
%Georgia Institute of Technology\\
%Atlanta, Georgia 30332--0250\\
%Email: mshell@ece.gatech.edu}
%\and
%\authorblockA{Twentieth Century Fox\\
%\and
%\authorblockN{James Kirk\\ and Montgomery Scott}
%\authorblockA{Starfleet Academy\\
%San Francisco, California 96678-2391\\

% avoiding spaces at the end of the author lines is not a problem with
% conference papers because we don't use \thanks or \IEEEmembership

%Homer Simpson\authorrefmark{2},
%Montgomery Scott\authorrefmark{3} and
%Eldon Tyrell\authorrefmark{4}}
%Telephone: (800) 555--1212, Fax: (888) 555--1212}
%\authorblockA{\authorrefmark{4}Tyrell Inc., 123 Replicant Street, Los Angeles, California 90210--4321}}

\maketitle

\begin{abstract}
% In the rapidly evolving field of robotic locomotion, it is critical to equip legged robots with dynamic and versatile movement capabilities.
% Conventional reinforcement learning techniques, though useful, often restrict robots to limited behaviors. 
% The complexity of navigating real-world environments necessitates a policy framework that supports a wide range of locomotion skills, ensuring flexibility and resilience. 
% To address the need for real-time locomotion capabilities in legged robots, we develop DiffuseLoco, a novel framework that employs diffusion models for training multi-skill locomotion policies from extensive offline datasets.
% Specifically, DiffuseLoco supports a broad range of locomotion skills and can be directly applied to real-world quadruped robots without the need for prior tuning, while achieving real-time performance on edge computing devices.
% Our evaluations demonstrate that DiffuseLoco outperforms state-of-the-art techniques in real-world scenarios, opening up new possibilities of leveraging imitation learning to create multi-skill controllers for legged locomotion.

This work introduces DiffuseLoco, a framework for training multi-skill diffusion-based policies for dynamic legged locomotion from offline datasets, enabling real-time control of diverse skills on robots in the real world. 
\changed{Offline learning at scale has led to breakthroughs in computer vision, natural language processing, and robotic manipulation domains.}
\changed{However, scaling up learning for legged robot locomotion, especially with multiple skills} in a single policy, presents significant challenges for prior online reinforcement learning methods.
\changed{To address this challenge, we propose a novel, scalable framework that leverages diffusion models to directly learn from offline multimodal datasets with a diverse set} of locomotion skills. 
With design choices tailored for real-time control in dynamical systems, including receding horizon control and delayed inputs, \method is capable of reproducing multimodality in performing various locomotion skills, zero-shot transfer to real quadrupedal robots, and it can be deployed on edge computing devices. Furthermore, \changed{\method demonstrates free transitions between skills and robustness against environmental variations.} 
Through extensive benchmarking in real-world experiments, \method exhibits better stability and velocity tracking performance compared to prior reinforcement learning and \changed{non-diffusion-based} behavior cloning baselines.
The design choices are validated via comprehensive ablation studies.
This work opens new possibilities for scaling up learning-based legged locomotion controllers \changed{through the scaling of large, expressive models and diverse offline datasets.}
\end{abstract}

\IEEEpeerreviewmaketitle

\section{Introduction}
\changed{Learning from large-scale offline datasets has led to breakthroughs in a large variety} of domains, such as computer vision and natural language processing, where scaling up both the size of models and datasets leads to improved performance and generalization~\cite{Kaplan2020Scaling, Sun2017Revisiting}. 
This has led to the development of powerful generative models, like diffusion models, which are able to model complex multi-modal data distributions~\cite{ryu2022pyramidal, rombach2022highresolution}\changed{, and generate high-quality images and videos.}

In robotics, learning from diverse offline datasets has also been shown to be an effective and scalable approach for developing more versatile policies for domains such as robotic manipulation \cite{brohan2023rt2, brohan2023rt1} and autonomous driving~\cite{chen2019deep, codevilla2018end, pan2017agile}. However, these domains typically involve agents that have low-dimensional action spaces (\textit{e.g.}, end-effector trajectory), with low re-planning frequency on inherently stable systems (\textit{e.g.}, robot arms or cars). For dynamical systems featuring higher degrees of freedom and more complex dynamics, such as legged robots, data-driven approaches have largely been focused on \changed{online reinforcement learning (RL) techniques~\citep{Hwangbo2019Learning, li2021reinforcement, Peng2020Learning}. Unlike offline learning, it can be difficult to scale online RL to both large models and datasets due to the requirements of online rollouts.} Most prior works have focused on dynamic motions with specialized single-skill models, and \changed{scaling towards a single model that can reproduce a diverse set of challenging locomotion skills remains an open problem.}

\changed{To this end, we present a novel approach that emphasizes learning agile legged locomotion skills at scale by solving two aforementioned challenges: offline learning from various data sources and the ability to learn a set of diverse skills.} We propose \emph{\method}, a framework that leverages expressive diffusion models to effectively learn the multi-modalities that exist in the \changed{diverse offline dataset without manual skill labeling.} Once trained, our controllers can execute robust locomotion skills on real-world legged robots for real-time control.

The primary contributions of this work include:
\begin{enumerate}
    \item A \changed{state-of-the-art multi-skill controller, leveraging expressive diffusion models, that learns agile bipedal walking and various quadrupedal locomotion skills within a single policy and can be deployed zero-shot on real-world quadrupedal robots.}
    \item A novel framework that directly learns from a diverse offline dataset for real-time control of legged robots, showing the benefits and potentials of offline learning \changed{at scale} for locomotion skills practically in a real-world scenario.
    \item Extensive real-world validation showing higher stability and lower velocity tracking errors compared to baselines, while demonstrating multi-modal behaviors with skill transitioning and robustness on terrains with varying ground frictions.

\end{enumerate}

This work opens up the possibility of leveraging \changed{large-scale} learning to create \changed{diverse and agile} multi-skill controllers for legged locomotion from offline datasets. For the first time, we show that it is feasible to zero-shot transfer \changed{such a diverse locomotion policy learned from a static dataset to real-world applications. This approach offers a scalable and versatile framework for learning-based control, allowing for continuous expansion of the dataset and integration of diverse skills from various data sources.} Codebase and checkpoints will be open-sourced upon the acceptance of this work. 
\label{sec:Introduction}

\section{Related Work}
Our proposed framework leverages diffusion models for \changed{multi-skill} legged locomotion control. In this section, we review the most closely related works on learning-based legged locomotion algorithms and applications of diffusion models in robotics. 

\subsection{\changed{Multi-skill Reinforcement Learning in Locomotion}}
\label{sec:related_multi}
 Recent advances in model-free RL have demonstrated promising results in developing agile locomotion skills for legged robots in the real world\changed{~\cite{Peng2020Learning, margolis2022rapid,rudin2022learning, li2023learning2, castillo2021robust, dao2022sim}}. \changed{Among them, prior works have demonstrated impressive performances on highly agile skills such as jumping, running and sharp turning on bipedal robots~\cite{li2024reinforcement, yu2022dynamic}, and walking on two feet with quadrupedal robots~\cite{fuchioka2023opt, smith2023learning, li2023learning}, which requires a high degree of agility and robustness with a floating-base robot.}
\changed{However, the majority of these agile locomotion skills are trained with single-skill RL and remain unscalable to large-scale learning of multiple locomotion skills.}

A simple and natural idea of learning multi-skill locomotion is to train separate policies for each skill, and then coordinate through extra high-level planning~\cite{huang2023creating, yang2020multi, hoeller2024anymal, kim2022humanconquad} \changed{or distill into a small-scale model via imitation learning such as DAgger~\cite{zhuang2023robot}. }Due to the inherent coordination difficulty \changed{and the requirements of online distillation, these methods remain unscalable to an increasing number of skills.} 

In comparison, learning multi-skill policies directly from scratch \changed{typically involves parameterized motions~\cite{Shao2022Learning, reske2021imitation} with limited sets of applicable motions, and more popularly, motion imitation methods through either reward shaping~\cite{zhang2024learning, klipfel2023learning, cheng2023extreme} or adversarial imitation learning~\cite{escontrela2022adversarial, li2023versatile, wu2023learning, yang2023generalized}.} \changed{However, this approach still faces challenges such as needing extra model-based trajectory optimization or well-trained expert policies for acquiring reference for agile motions and the limited expressiveness of simple models in online RL frameworks in learning diverse skills.}

In general, learning a diverse, agile multi-skill policy with online RL remains challenging.
\changed{For example, while existing RL methods have successfully combined skills like jumping and trotting~\cite{zhang2024learning, yang2023generalized}, quadrupedal walking and standing on hind legs with wheels without bipedal walking}~\cite{vollenweider2023advanced}, \changed{a combination of more diverse and agile skills such as stable bipedal walking and quadrupedal hopping in a single policy has not yet been demonstrated.}

\subsection{\changed{Offline Learning in Locomotion Control}}
{Compared to online learning, offline learning offers better scalability, a simpler training scheme, and an effective way to re-use data, yet prior works in learning low-level locomotion control from offline datasets remain limited.}
% In contrast, there have been limited attempts to learn low-level locomotion control from offline datasets. 
Most prior works focus on simple simulated tasks, such as Gym locomotion tasks, \changed{with behavior cloning (BC)~\cite{torabi2018behavioral}} and offline RL~\cite{levine2020offline}. 
Among them, some leverage Q-learning on offline datasets~\cite{nakamoto2023cal, kumar2020conservative, kostrikov2021offline}, or supervised learning techniques\changed{~\cite{wang2023diffusionexpressive, chen2021decision, xu2022policy, janner2021offline}}. 
However, these tasks are oversimplified and do not adequately consider the complexities in real-world scenarios. 

An alternative is the use of offline data as a foundation for online learning~\cite{nair2018overcoming, vezzi2023two}. Among them, \citeauthor{smith2023learning} develops baseline policies from offline datasets to bootstrap online learning on real robots.
Yet, this approach still requires online learning. 

\changed{Another recent work develops offline learning on humanoid locomotion with real robots~\cite{radosavovic2024humanoid}, with the scope limited to only one walking skill.}
In comparison, \changed{the efficacy of learning completely from offline datasets, especially at a larger scale than a few simple skills remains unproven in legged locomotion control}.

\subsection{Diffusion Models in Robotics}

Recent advances have seen increasing applications of diffusion models in control and planning systems. Some prior works integrate them into the learning pipeline, including the discriminators in adversarial IL for legged control~\cite{wang2023diffail}, and reward models for RL~\cite{nuti2023extracting, wang2023diffusion}. However, the use of small multi-layer perceptron (MLP) networks as policies presents similar challenges in exploring diverse, multi-skill learning~\cite{wang2023diffusionexpressive}. Diffusion models are also employed in high-level trajectory planning~\cite{janner2022planning, huang2023diffusion}, safe planning~\cite{xiao2023safediffuser}, and goal generation for low-level controllers~\cite{kapelyukh2023dall, ajay2022conditional}, as well as enhancing visuomotor planning for manipulation tasks~\cite{pearce2023imitating, reuss2023goalconditioned, mishra2023generative}. However, most prior works are limited to simulation environments only. 

Emerging efforts to apply diffusion models to real-world robot manipulation include using diffusion to manage a variety of manipulation tasks with visual inputs and incorporating self-supervised learning and language conditioning~\cite{chi2023diffusion, li2024crossway, chen2023playfusion, ha2023scaling}. 
\citeauthor{yoneda2023noise} leverages the reverse diffusion process for shared autonomy with a human user in end-effector planning.
Additionally, hierarchical frameworks are being developed to handle tasks requiring multiple skills, pushing towards generalist policies~\cite{black2023zero, xian2023chaineddiffuser, octo_2023}. However, these prior works primarily focus on high-level planning on manipulation systems, featuring a low-dimensional action space (\textit{e.g.}, end-effector position), low-replanning frequency (\textit{e.g.}, around 10~\unit{Hz}), and inherently more stable dynamics. 

In contrast, using diffusion models for high-frequency, low-level control remains limited~\cite{chen2023score}. The most relevant work uses online RL to train diffusion-based actor policies in simpler simulation settings~\cite{yang2023policy}, but transitioning to real-world applications with high-frequency feedback control presents significant challenges due to the instability and rapid dynamics of legged robots~\cite{westervelt2018feedback}. This work aims to leverage diffusion models in low-level control for legged locomotion \changed{and demonstrate the advantages of multimodality and scalability in the real world.}
\label{sec:Review}

\section{Versatile Framework for Diffusion Locomotion Control From Offline Dataset}\label{sec:overview}

In this section, we provide an overview of \textit{\method}, a framework designed to generate and utilize offline datasets for training multi-skill locomotion policies \changed{scalably}. 
\method is based on the diffusion model and is designed to train a low-level multi-skill locomotion policy from offline datasets containing \changed{multiple agile locomotion skills with} diverse behaviors. 
\method ~\changed{represents the state-of-the-art performance in multi-skill locomotion, as it is able to learn both bipedal and quadrupedal locomotion skills in a unified policy, and can be deployed robustly zero-shot on hardware.}
\method ~\changed{addresses} several challenges inherent to learning from offline data, including generating diverse skills with the same goals, handling data variability, and ensuring real-time control in real-world environments.

\begin{figure}[t]
\centering
\includegraphics[width=\linewidth]{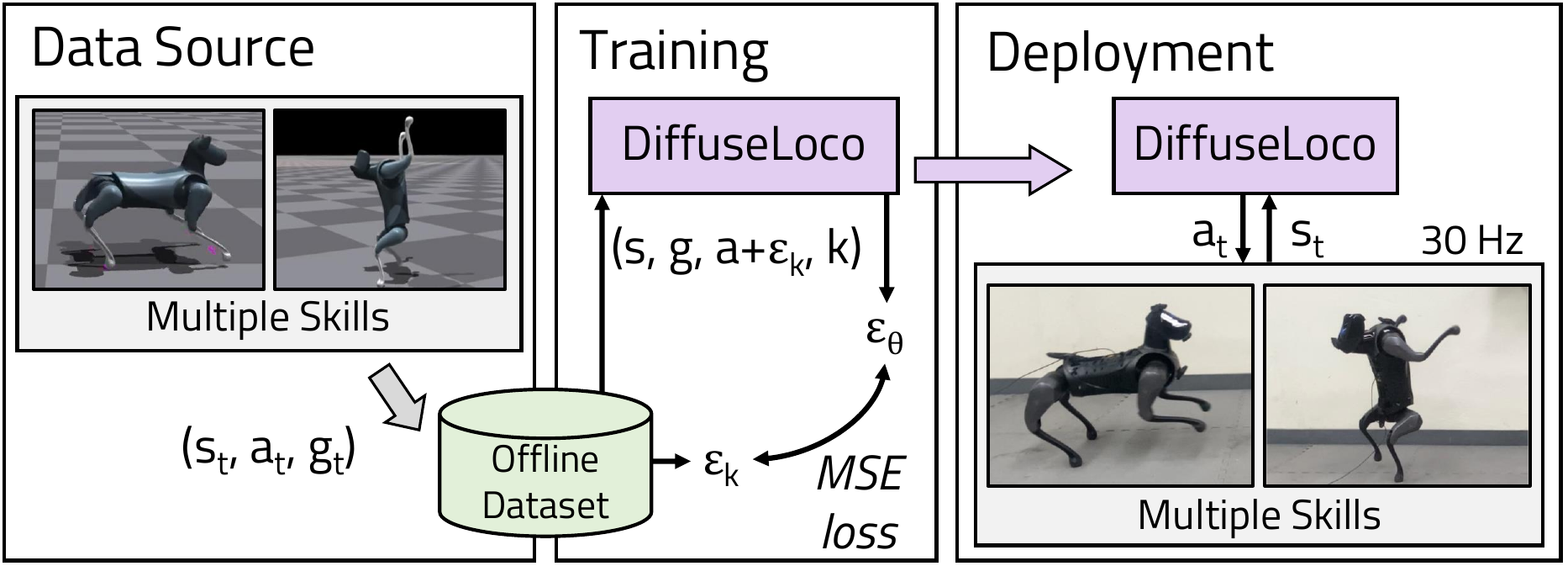}
\refstepcounter{figure}
\caption{\label{fig:workflow-diagram} Overview of the three stages of \method. First, we generate or utilize an offline dataset with demonstrations of a set of skills gathered with different methods (left). Then, we train \method policy with DDPM loss on trajectories within the dataset (middle). Finally, \method policy is deployed on robots in the real world and executes a diverse set of agile skills (right). }
\end{figure}

\begin{algorithm}[t]
\caption{\method Algorithm}\label{algo:workflow-algorithm}
\begin{algorithmic}[1]
\State Initilize: N source policies $\pi_\text{src}^1 \dots, \pi_\text{src}^N$, Empty \changed{or Existing} Offline Dataset $\mathcal{D}_\text{src}$, Diffusion Model $\pi_{\theta}$
\State // OBTAIN DATA FROM $\pi_\text{src}$
\Repeat
\State Sample $n$ uniformly from $\{1, \dots, N$\}
\State Sample environment dynamics $p(\mathbf{s}'|\mathbf{s},\mathbf{a})$
\State \changed{Reset source-specific state $\mathbf{s}_{t}^\text{src}$} 
\For{$t = 1$ \textbf{to} $T$}
    \If{$t \mod \text{random\_goal\_step} = 0$}
        \State Sample goal $\mathbf{g}$
    \EndIf
    \State \changed{Act source policy $\mathbf{a}_t^\text{src} = \pi_\text{src}^n(\mathbf{s}_t^\text{src},\mathbf{g}_{t})$}
    \State \changed{Record source-agnostic state $\mathbf{s}_{t}$}
    \State \changed{Step environment ${\mathbf{s}_t^\text{src}}' = $ environment.step($\mathbf{a}_t^\text{src}$)}
    \State \changed{Record source-agnostic action $\mathbf{a}_{t}$}
    \State Add data to offline dataset $\mathcal{D}_\text{src} \leftarrow (\mathbf{s}_{t}, \mathbf{a}_{t}, \mathbf{g}_{t})$
\EndFor
\Until{desired}
\State // TRAIN ON OFFLINE DATASET
\For{each epoch}
    \For{each $(\mathbf{s}_\text{traj}, \mathbf{a}_\text{traj}, \mathbf{g}_\text{traj})$ in $\mathcal{D}_\text{src}$}
        \State Compute the loss $L(\theta)$ with Eqn. \ref{eqn:ddpm-loss-goal-delay}
        \State Update model parameters $\theta$ to minimize loss $L(\theta)$
    \EndFor
\EndFor
\end{algorithmic}
\end{algorithm}

A schematic illustration of the \method framework is shown in Figure~\ref{fig:workflow-diagram}. Our framework consists of three stages: 

\paragraph{Data Source} We start with collecting \changed{a new or utilizing an existing offline} dataset consisting of multiple skills.
To generate or expand a dataset, we first obtain single-skill control policies as the source policies. 
These policies are conditioned on given goals $\mathbf{g}$ (commands), such as velocity commands and base heights, that are skill-agnostic.  
\changed{Note that the source policies can be obtained with different methods. With the assumption of the frequency of low-level control being the same, the observation and action spaces of the \emph{source} policies can be vastly different. 
Thus, we need to collect a set of \emph{source-agnostic} state and action pairs across all source policies. For legged robots, the widely-used source-agnostic states and actions are the proprioceptive feedback $\mathbf{s}_t$ directly from the robot and the joint-level PD targets $\mathbf{a}_t$ directly to the robot, respectively.}
As illustrated in Algo.~\ref{algo:workflow-algorithm}, we start an episode where the robot is controlled by the $i^{\text{th}}$ source policy $\pi^i$. \changed{The policy acts at a source-specific state $\mathbf{s}_t^\text{src}$, and only the source-agnostic state-action-goal pairs ($\mathbf{s}_t$, $\mathbf{a}_t$, $\mathbf{g}_t$) are collected} during the rollout of the robot's closed-loop dynamics until it reaches the maximum episode length $T$. 
In addition, the goal $\mathbf{g}_t$ will be re-sampled within the command range after a time interval within the episode. \changed{In this work, we leverage cheap simulation data as an example, but since \method effectively re-uses data, we can scalably extend to more expensive data collection process, potentially from real-world hardware. The details of dataset generation in our experiments can be found in the Appendix \ref{appx:dataset}.}

\paragraph{Training} In the second step, we train our \method policy from the \changed{offline} dataset in an end-to-end manner. Let input state and goal history length be $h$ and output action prediction length be $n$. During training, we sample a segment of state trajectory $\mathbf{s}_\text{traj}$ and corresponding action and goal sequences, $\mathbf{a}_\text{traj}$ and $\mathbf{g}_\text{traj}$. 
We sample a diffusion step $k$ randomly from $\{1, \dots, K\}$, and sample a Gaussian noise $\epsilon_k$ to add to the action sequence. 
Then, a transformer-based denoising model takes the noisy action sequence along with states trajectory $\mathbf{s}_\text{traj}$, goal trajectory $\mathbf{g}_\text{traj}$, and diffusion step $k$ as input, and predicts the added noise as $\epsilon_\theta$. 
The predicted noise $\epsilon_\theta$ is then regressed to match the true noise $\epsilon_k$ with mean square error loss. 
In this way, the denoising model is learned to generate sequences of low-level actions conditioned on robot states and goals from the dataset. 
Details of the model architecture and training objective are introduced in Sec.~\ref{sec:diffusion-for-realtime}.

\paragraph{Deployment} In the last stage, we zero-shot transfer the trained DiffuseLoco policy on the robot hardware.
During deployment, the DiffuseLoco policy takes a sequence of pure noise sampled from a Gaussian distribution and denoises it conditioned on the state trajectory $\mathbf{s}_\text{traj}$ from the robot hardware and the given goal $\mathbf{g}_\text{traj}$. 
The denoising process is repeated for $K$ iterations to generate a sequence of actions, but only the \textit{immediate} action $\mathbf{a}_t$ is used as the robot's joint-level PD targets. 
After executing this action, the \method policy takes a new sequence of states from the robot and updates the immediate action from the newly generated action sequence. 
This is designed to align with the Receding Horizon Control (RHC) framework, instead of interpolating the action sequence at high frequency as previously used by other diffusion-based work~\cite{kapelyukh2023dall, li2024crossway}. \changed{RHC enables \method to replan rapidly with fast-changing states of the robot to ensure up-to-date actions while keeping future steps in account.} 
However, since the diffusion model has a large number of parameters \changed{and the diffusion process involves multiple denoising steps}, we must accelerate the inference to be faster than the control frequency to achieve this RHC manner. \changed{The acceleration techniques are detailed in Appendix \ref{appx:acceleration}}, which ultimately enables running the \method policy on an edge-compute device that can be mounted on the robot.

\section{Diffusion Model for Real-Time Control}\label{sec:diffusion-for-realtime}

\begin{figure}[t]
\centering
\includegraphics[width=\linewidth]{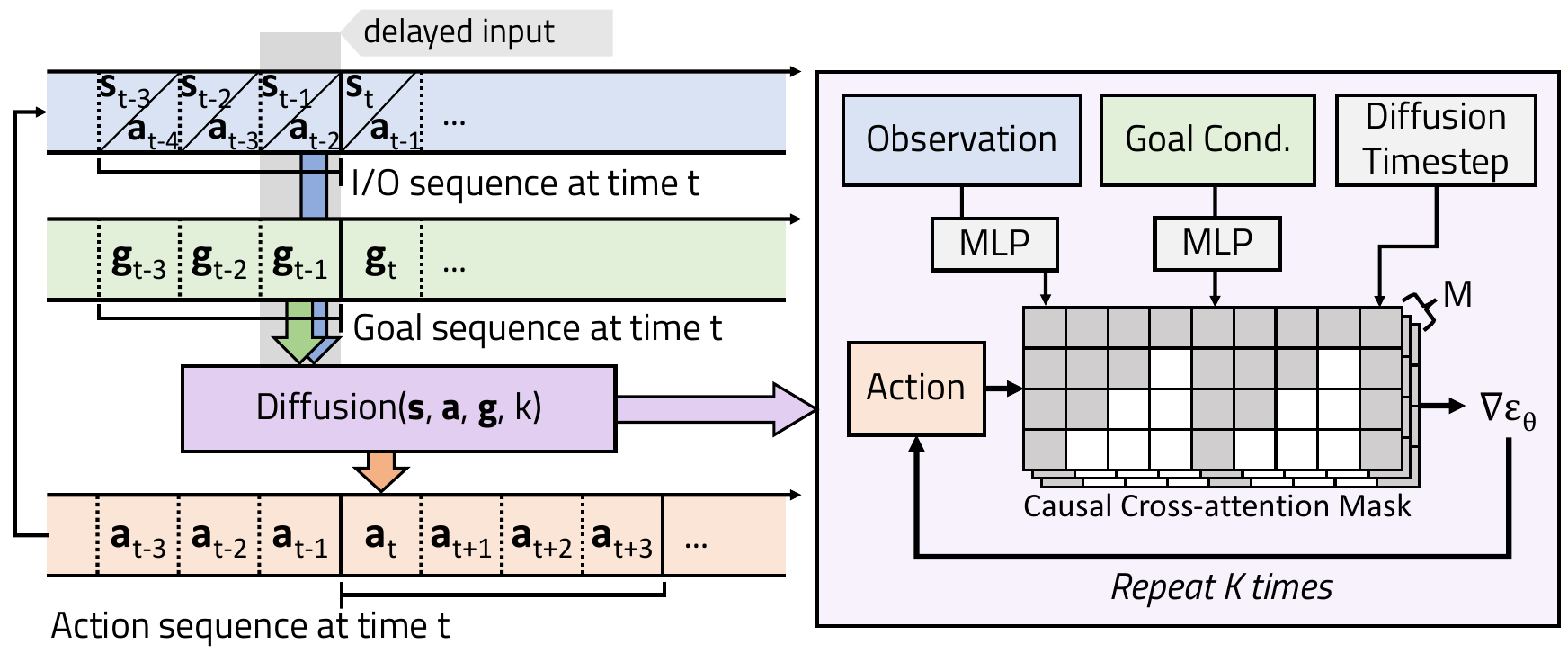}
\caption{\label{fig:system-diagram} 
The \method architecture. At time step $t$, it takes in a delayed $h$-step history of proprioceptive states $\mathbf{s}_{t-h-1:t-1}$, goals $\mathbf{g}_{t-h-1:t-1}$, and actions $\mathbf{a}_{t-h-2:t-2}$, and predicts a sequence of $n$ future actions $\mathbf{a}_{t:t+n}$ for the robot's actuators. 
First, separate MLP encoders map state and goal into embeddings which, with a one-hot diffusion step, are queried by noisy action tokens via $M$ transformer decoder layers for denoising. After $K$ denoising iterations, the predicted action sequence is generated and we feed the executed action back to the model’s input. 
The model is trained through end-to-end imitation learning.
}
\end{figure}

Having introduced the framework of \method, we now begin to develop its backbone: a diffusion model for locomotion control, shown in Fig.~\ref{fig:system-diagram}, with a special focus on design choices for real-time control and inference acceleration. 

\subsection{DDPM for Control}
To model multi-modal behaviors from diverse datasets, we leverage Denoising Diffusion Probabilistic Models (DDPM)~\cite{ho2020denoising} \changed{with a transformer backbone} to model different skills that can be applied to achieve a common goal. DDPM is a class of generative models in which the generative process is modeled as a denoising procedure, often referred to as Stochastic Langevin Dynamics, \changed{expressed in the following equation, 
\begin{equation}\label{eqn:ddpm-original}
\mathbf{x}^{k-1} = \alpha ~\left(\mathbf{x}^k - \gamma \epsilon_\theta(\mathbf{x}^k, k) + \mathcal{N}(0, \sigma^2 I)\right)
\end{equation}
where} $\mathcal{N}(0, \sigma^2 I) $ denotes the sampled noise from a DDPM scheduler, $\alpha$, $\gamma$, and $\sigma$ are its hyperparameters: $\alpha$ regulates the rate at which noise is added at each step, $\gamma$ represents the denoising strength, and $\sigma$ defines the noise level. 
For clarity, we now use subscripts $\mathbf{}_{t-a:t-b}$ to indicate trajectories from timestep $t-a$ to $t-b$, replacing subscripts $\mathbf{}_\text{traj}$.
To generate the action trajectory for control, an initial noisy action sequence, $\mathbf{a}_{t:t+n}^K$, is sampled from Gaussian noise, and the DDPM conditioned on states $\mathbf{s}_{t-h:t}$, goals $\mathbf{g}_{t-h:t}$, and previous actions $\mathbf{a}_{t-h-1:t-1}$ undergoes $K$ iterations of denoising steps. 

\changed{Unlike previous works applying DDPM in manipulation~\cite{chi2023diffusion}, the inclusion of previous actions, \textit{i.e.}, I/O history, helps the policy to better perform system identification and state estimations for legged locomotion control, as evaluated in \cite{li2024reinforcement}.}
\changed{Furthermore, instead of concatenating state and goal into a single embedding~\cite{chi2023diffusion, li2024crossway}, we conveniently leverage the transformer's attention mechanism to assign different attention weights to separately embedded robot's I/O and static goals, enabling the policy to adjust focus between adapting to dynamic environments and achieving goals. We find that these modifications result in better command tracking performance and robustness, as shown in  Sec.~\ref{sec:ablation-study}.}

The denoising process yields a sequence of intermediate actions characterized by progressively decreasing noise levels: $\mathbf{a}^K, \mathbf{a}^{K-1}, \ldots, \mathbf{a}^0$, until the desired noise-free output, $\mathbf{a}^0$, is attained. 
This process can be expressed as the following equation:
\begin{equation}\label{eqn:ddpm-denoise-goal}
\begin{split}
\mathbf{a}^{k-1}_{t:t+n} = \alpha ~ &\left(\mathbf{a}^k_{t:t+n} - \gamma \epsilon_\theta(\mathbf{a}^k_{t-h-1:t+n}, \mathbf{s}_{t-h:t}, \mathbf{g}_{t-h:t}, k) \right. \\
& \left. + \mathcal{N}(0, \sigma^2 I)\right)
\end{split}
\end{equation}\noindent
where $\mathbf{a}^{k}_{t:t+n}$ represents the output at the $k^{\text{th}}$ iteration, and $\mathbf{\epsilon}_\theta (\mathbf{a}^k_{t-h-1:t+n}, \mathbf{s}_{t-h:t}, k)$ represents the predicted noise from the denoising model $\epsilon_\theta$, which is parameterized by $\theta$, with respect to $\mathbf{a}^k_{t-h-1:t+n}$, $\mathbf{s}_{t-h-1:t}$, and iteration $k$. 

During training, we opt to use the simplified training objective proposed by \citet{ho2020denoising},
\begin{equation}\label{eqn:ddpm-loss-original}
l = MSE \left( \epsilon_k, \epsilon_{\theta}(\mathbf{a}_{t-h-1:t+n} + \epsilon_k, \mathbf{s}_{t-h:t}, \mathbf{g}_{t-h:t}, k) \right).
\end{equation}\noindent
where $\epsilon_k$ is the sampled noise at iteration $k$. \changed{The detailed model architecture can be found in Appendix \ref{appx:architecture}.}

\subsection{Delayed Input and Predicted Actions}
To achieve real-time deployment, we introduce the technique of predicting current actions using delayed inputs, overcoming the challenge posed by the extensive inference times of large models like transformers, \changed{which exceed the common control frequency requirement of 30-50Hz.}

In the \method policy, we use delayed inputs from the previous timestep—$\mathbf{s}_{t-h-1:t-1}$, $\mathbf{a}_{t-h-2:t-2}$, and $\mathbf{g}_{t-h-1:t-1}$—to predict actions for the current timestep $\mathbf{a}_{t:t+n}$. By initiating inference for the next action before the current state $\mathbf{s}_{t}$ is received, we can run the inference in parallel and ensure actions are the most up-to-date. This design choice is favored for two reasons: 
\changed{First, we train a sequence of predictions, instead of one-step prediction autoregressively, which is suitable for generating actions further in time than the nearest timestep.}
Second, these larger-scale models are adept at handling higher input delays. \changed{Since delays (zero-order-hold) exponentially deteriorate the stability of the system, this marks an advantage over small-scale MLP policies, which typically manage delays less than one control step, for instance, 25\% less than \method as noted in~\cite[Table IV]{li2024reinforcement}. }

In this way, we modify the denoising process in Eqn.~\ref{eqn:ddpm-denoise-goal} to the following, 
\begin{equation}\label{eqn:ddpm-denoise-goal-delay}
\begin{split}
\mathbf{a}^{k-1}_{t:t+n} = \alpha  ~&\left(\mathbf{a}^k_{t:t+n}\right. \\
& \left.- \gamma \epsilon_\theta(\mathbf{a}^k_{t-h-2:t+n}, \mathbf{s}_{t-h-1:t-1}, \mathbf{g}_{t-h-1:t-1}, k) \right.\\
&+ \left. \mathcal{N}(0, \sigma^2 I) \right)
\end{split}
\end{equation}
and the loss function in Eqn.~\ref{eqn:ddpm-loss-original} to the following, 
\begin{equation}\label{eqn:ddpm-loss-goal-delay}
\begin{split}
l = MSE \left(\epsilon_k, \epsilon_{\theta}( \right. &\mathbf{a}_{t-h-2:t+n} + \epsilon_k,\\
& \left. \mathbf{s}_{t-h-1:t-1}, \mathbf{g}_{t-h-1:t-1}, k) \right)
\end{split}
\end{equation}

Note that for more complex tasks, larger models may require more time for inference, then our method can be extended by delaying more than one timestep of inputs. 
However, more delayed steps will result in more challenging learning complexity because of the lack of recent state feedback. 

\begin{remark}
    Another possible solution to run inference before $\mathbf{s}_{t}$ arrives is to predict $\mathbf{s}_t$ given $h$-step history of previous states, $\mathbf{s}_{t-h:t-1}$ with either a model-based Kalman Filter~\cite{4282823} or a learning-based prediction model~\cite{ji2022concurrent}. However, state prediction inevitably introduces extra prediction error or bias to the policy. 
\end{remark}

\subsection{\changed{Sampling Techniques}}\label{subsec:ddim}
To accelerate diffusion models, especially during robotic deployment, prior work often uses samplers like the Denoising Diffusion Implicit Models (DDIM)\cite{song2022denoising}, which employ a deterministic process to reduce sampling steps and speed up inference, albeit with some loss in sample quality. \changed{However, we find that DDIM is less suited for real-time control on legged robots due to its less accurate action outputs, which increases the compounding error of each step that leads to more frequent out-of-distribution scenarios from training to real-world deployment.} Consequently, in \method, we continue using the Denoising Diffusion Probabilistic Model (DDPM), maintaining the same number of denoising iterations as during training. Our detailed comparisons in Sec.\ref{exp:ddpmvsddim} show that DDPM enhances robustness and performance over DDIM in real-time control scenarios.

% ========= changes tracked here (2024-04-19) ===========
% == (not including captions and figures)

\label{sec:Method}

\section{\changed{Results: Model Capacities}}
\label{sec:multi-skill}
\changed{
To scale up learning locomotion skills as discussed in Sec. \ref{sec:Introduction}, the critical questions we need to address are whether \method can a) be trained with various sources of demonstrations and b) incorporate a diverse set of skills present in the dataset.
In this work, we answer these questions by presenting a state-of-the-art five-skill policy that combines four quadrupedal skills, and more importantly, a bipedal locomotion skill, for quadrupedal robots. As discussed in Sec. \ref{sec:related_multi}, the ability to learn these diverse skills with a single model, including an agile bipedal locomotion policy along with quadrupedal skills, remains challenging and has not yet been demonstrated by prior RL frameworks. 
}
\subsection{\changed{Learning from Diverse Data Sources}}
\begin{figure}[t]
\centering
\includegraphics[width=\linewidth]{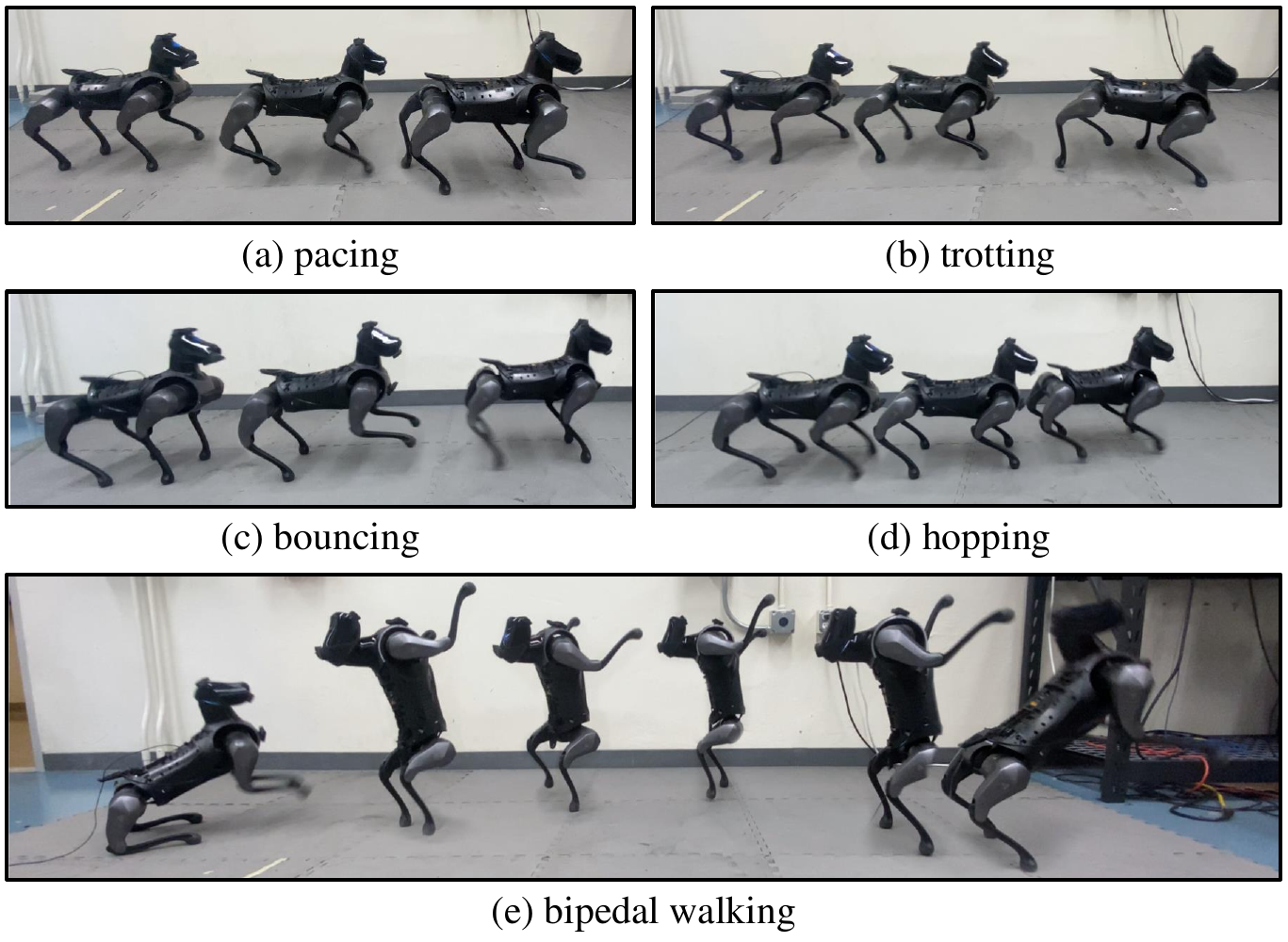}
\caption{\label{fig:skills} 
\changed{Snapshots of five diverse agile locomotion skills with the \method policy. 
This represents a leading effort in developing a single policy that can combine an agile bipedal walking skill with other quadrupedal skills and can be deployed on real-world robots. }
}
\end{figure}

\changed{
Fig. \ref{fig:skills} illustrates the five skills acquired by \method for a quadrupedal robot, encompassing quadrupedal walking skills like trotting and pacing, jumping skills such as hopping and bouncing, and an agile bipedal walking skill. Among these, trotting and pacing are trained using AMP~\cite{escontrela2022adversarial}, hopping and bouncing through nominal CPG curves~\cite{Shao2022Learning}, and bipedal locomotion with symmetry augmentation~\cite{su2024leveraging}. After collecting demonstrations of these skills separately in simulation, we directly learn from this combined dataset and achieve robust zero-shot transfer to actual hardware. This capability surpasses previous offline learning methods that were largely confined to simulated environments with simplified dynamics, showcasing \method's robust real-world performance.
}

\changed{
The ability to learn from diverse skill sources is crucial for scaling locomotion learning, especially for this five-skill controller. For instance, bipedal locomotion requires a distinct framework with specific early termination conditions and reward landscape compared to quadrupedal skills. With these specialized conditions, basic symmetry augmentation can already yield effective stepping patterns, whereas, learning the same skill with prior multi-skill RL approaches like AMP~\cite{Peng2020Learning} or motion imitation involves complex reference motions and trajectory optimization~\cite{vollenweider2023advanced}, representing a significant engineering challenge. Despite the different requirements for observation, action spaces, and auxiliary signals such as phase signals across various RL methods used in this work, \method simplifies the process by solely relying on basic proprioceptive inputs to accomplish all skills, thanks to the powerful multimodal capability of diffusion models and receding horizon control.
}

\subsection{Skill Transitioning}

\changed{
More importantly, we demonstrate \method's capacity to transition freely between skills, 
which are not originally present in the dataset, such as transitioning from hopping to bipedal walking and then to pacing, as shown in Fig. 1. This sequence highlights the robustness of the \method policy against variations in starting state and the stability required to execute these skills successfully. Out of five consecutive runs, we do not experience any failure. Additional examples of skill transitions are available in Appendix \ref{appx:transitioning}.
}

\begin{figure*}[t]
\centering
\includegraphics[width=0.99\textwidth]{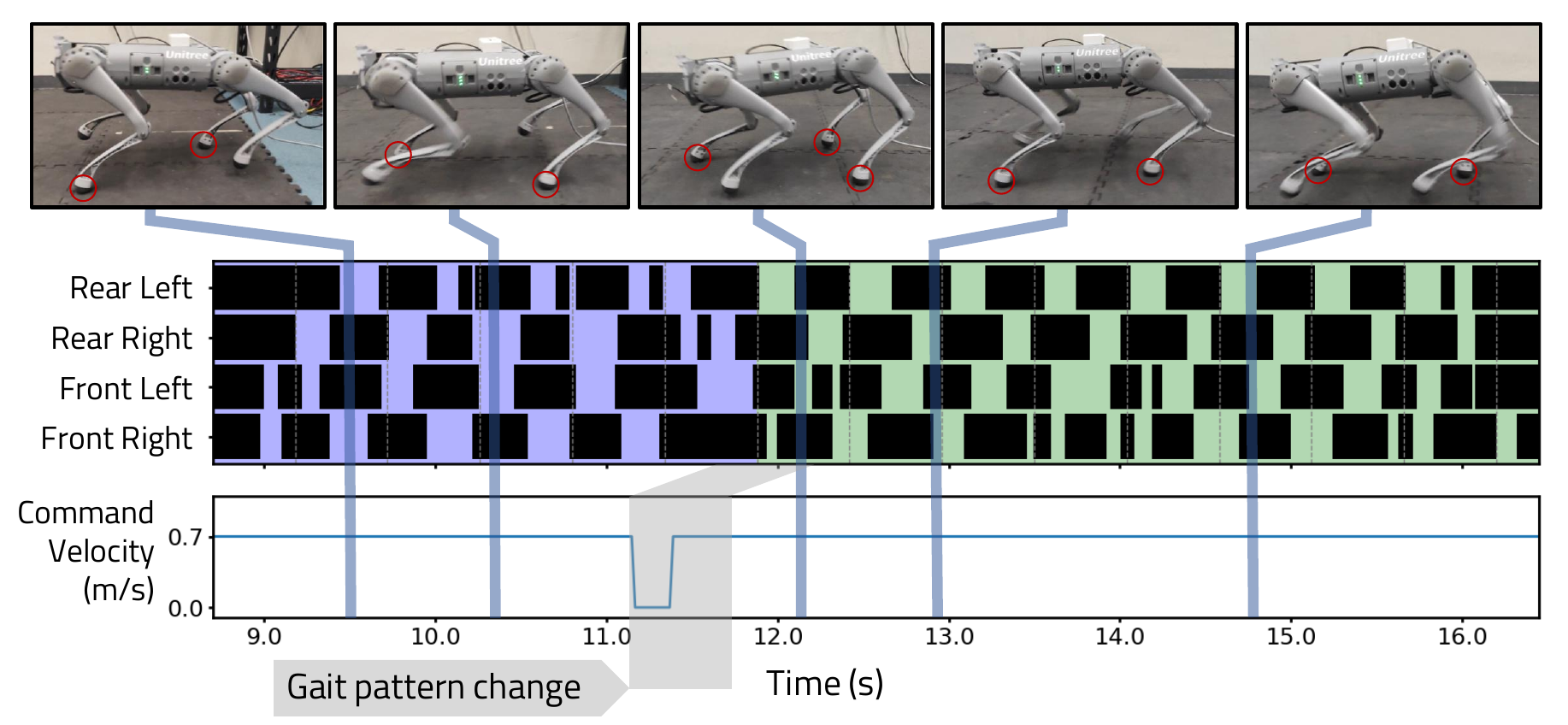}
\caption{\label{fig:trotting-mode} Foot contact map indicating stable walking and skill switching with \method policy and velocity commands. The red circle denotes the legs that are in contact with the ground. The robot initially walks using trotting skill, indicated by a purple background, then switches to pacing, shown in green, following a command change that involves a sudden stop and resume. 
We emphasize \method's ability to maintain different modalities for stable walking under the same command, switching modalities only when necessary.
} 
% Upon receiving a stop command, the robot reduces speed by grounding three feet, coincidentally placing it in a posture such that it seamlessly continues to walk with the pacing skill upon command resumption. }
\end{figure*}

In addition to transitions under \emph{different} goals (commands), the \method policy also demonstrates the ability to perform both trotting and pacing under the \emph{same} goal. In our experiment, as shown in Fig. \ref{fig:trotting-mode}, the policy begins with trotting and only switches to pacing when a sudden braking event significantly alters the contact sequence. 
\changed{This highlights 
\method's effectiveness in learning and adhering to different modes from the offline dataset, committing to a single mode within each rollout unless prompted by external disturbances.}

\subsection{Extension to Bipedal Robots}
\label{sec:biped}
In addition to quadrupedal robots, we also demonstrate the effectiveness of our method on high-dimensional, highly non-linear bipedal robots with a human-sized Cassie in the MuJoCo simulation. \changed{First, we collect demonstrations evenly from two separately trained single-skill RL policies on walking and running, adapted from \cite{li2024reinforcement}. After training directly on this aggregated dataset, our method successfully learns both skills within a single policy. Furthermore, as shown in Fig. 1, our policy can transition from walking to running smoothly without specific transition data in the dataset, in addition to maintaining each skill's stability before and after transitions. 
% This also marks a significant advancement over existing RL frameworks, which have not previously achieved such diverse multi-skill locomotion in bipedal robots.
This demonstrates one of the initial working combinations of these skills on bipedal robots.
}

\begin{figure}[t]
  \centering
  \subfloat[Turf]{\includegraphics[width=0.268\textwidth]{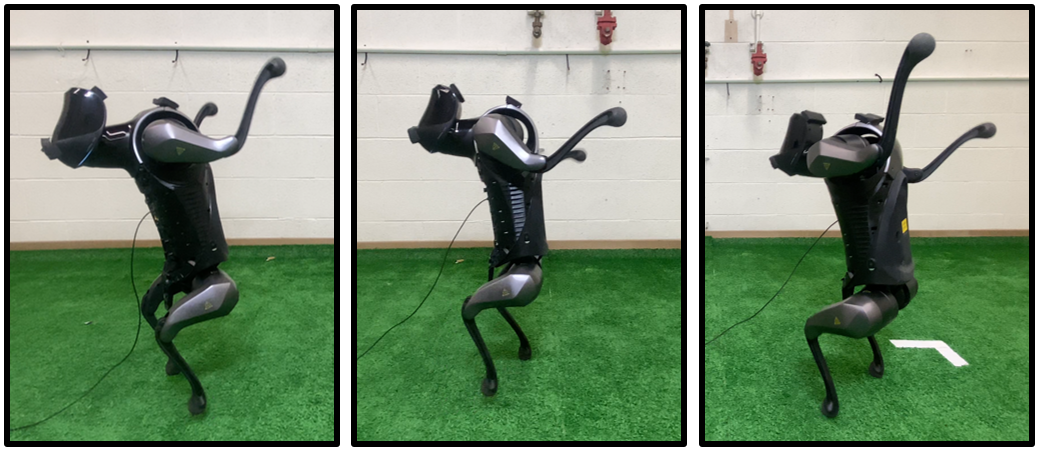}\label{fig:robustness-stand-grass}}
  \hfill
  \subfloat[Turf]{\includegraphics[width=0.205\textwidth]{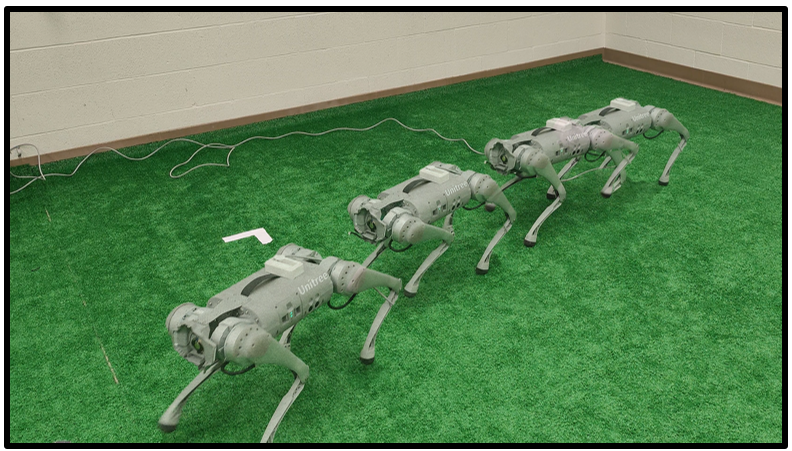}\label{fig:exp-robustness-grass}}
  \hfill
  
  \subfloat[Bare Floor]{\includegraphics[width=0.268\textwidth]{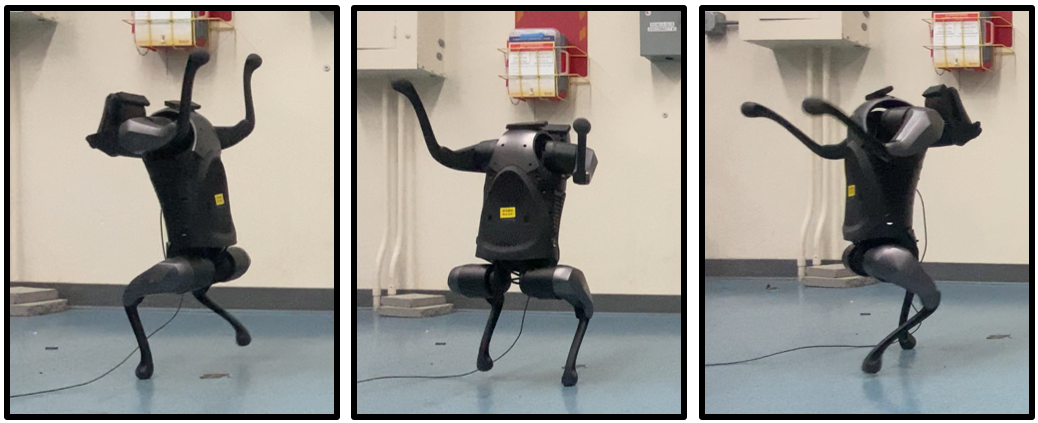}\label{fig:robustness-stand-bare}}
  \hfill
  \subfloat[Bare Floor]{\includegraphics[width=0.205\textwidth]{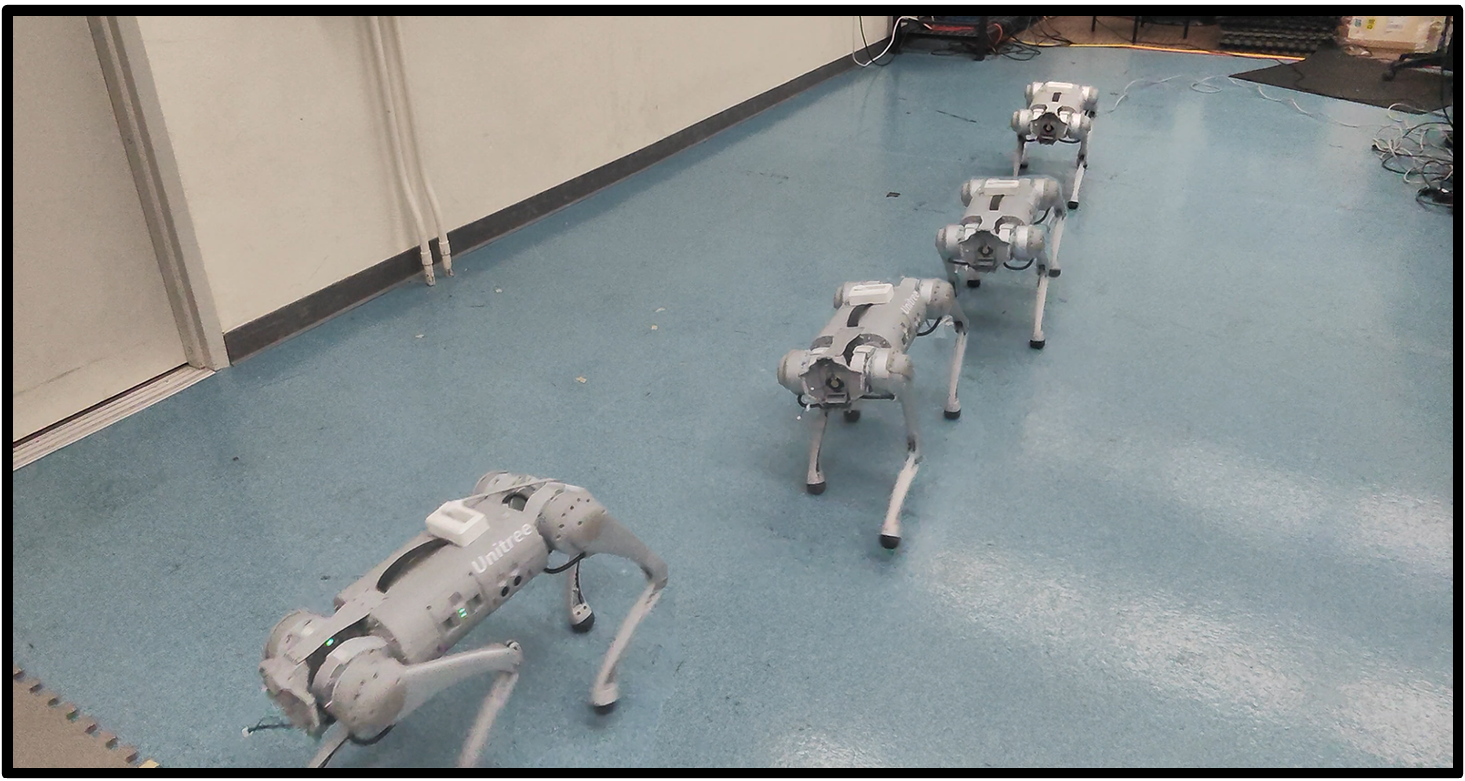}\label{fig:exp-robustness-floor}}
  \hfill
  
  \subfloat[Half Padded Floor]{\includegraphics[width=0.268\textwidth]{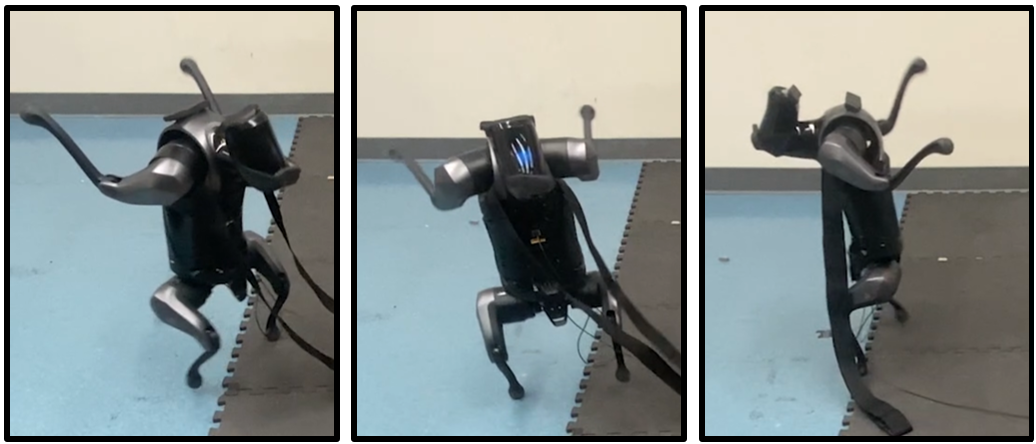}\label{fig:robustness-stand-half}}
  \hfill
  \subfloat[Over a Wooden Step]{\includegraphics[width=0.205\textwidth]{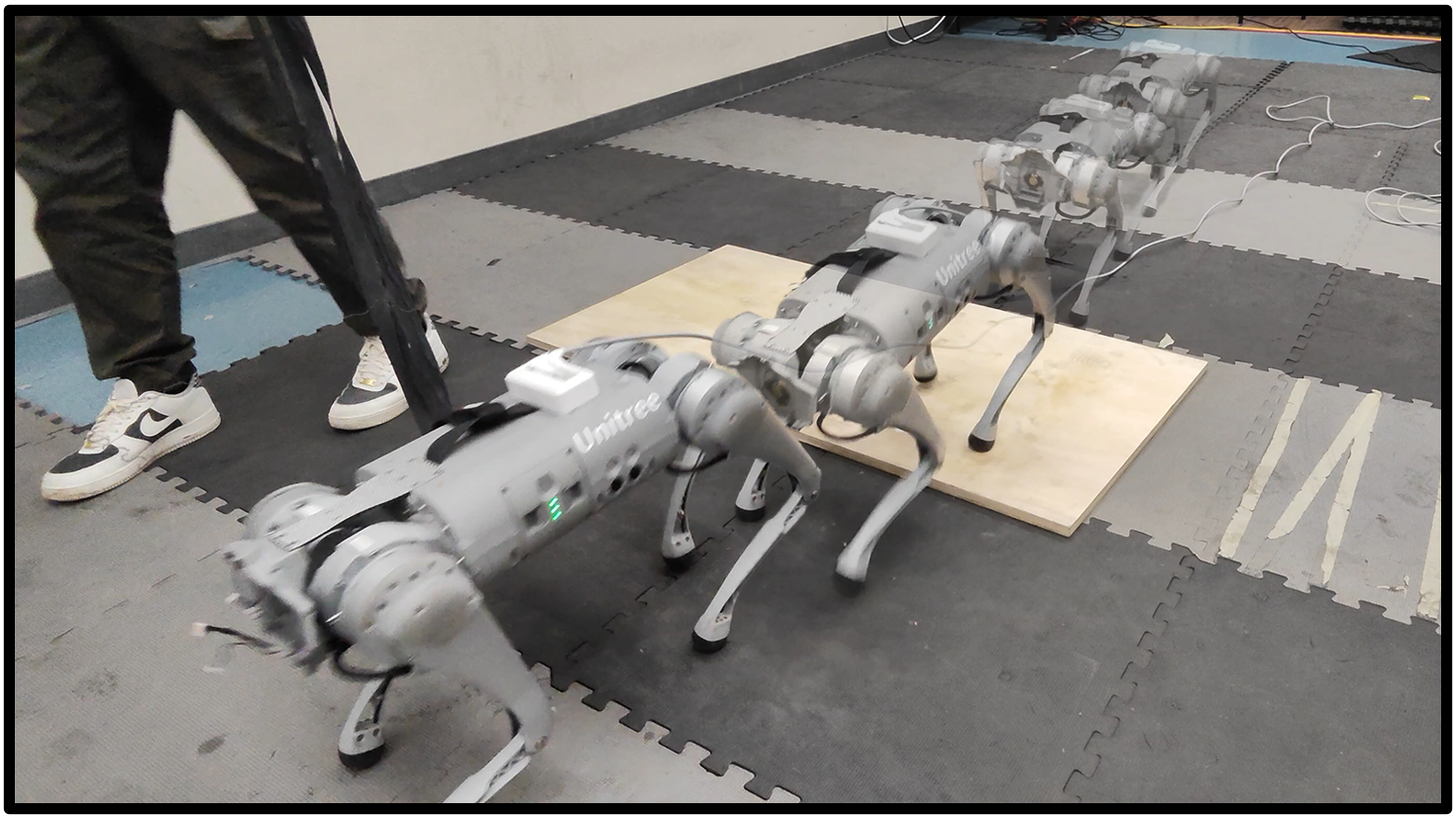}\label{fig:exp-robustness-plank}}
  \caption{Depiction of \method's robustness on different ground conditions and terrains: \changed{bipedal walking on (a) turf terrain, (c) vinyl composite floor, and (e) half padded floor, where the ground heights, friction and restitution forces on the two standing legs are different;} quadrupedal walking on (b) turf, (d) bare floor, (f) over a thick wooden board as a variation in the terrain height.} 
  \label{fig:exp-robustness}
\end{figure}
\subsection{Robustness}
To demonstrate \method's robustness, we show both quadrupedal and bipedal locomotion over different ground conditions, including padded floor, bare floor, turf, and over small terrain variations, as shown in Fig. \ref{fig:exp-robustness}. 
\changed{We especially emphasize bipedal walking over the half-padded floor, demonstrating a high degree of robustness to the differences in ground heights, as well as friction and restitution forces on each side of the two standing legs.} 

\changed{
Admittedly, \method's robustness is subpar compared to the state-of-the-art single-skill RL policy. However, while the RL policy is mostly limited by simulation randomization, one of the most desirable advantages of \method is its ability to learn scalably on offline real-world data without accessing expert policies in training or repeated real-world collection processes. 
More importantly, as we will show in Sec. \ref{exp:norand}, \method shows the potential to improve its robustness by absorbing data from various dynamics. By expanding our dataset with more diverse, potentially real-world data, we anticipate \method to achieve better robustness progressively.
}

\changed{
In conclusion, in addressing question a), we have demonstrated that \method is invariant to the source of offline demonstrations, and can be trained with data from multiple specialized RL frameworks. More importantly, \method shows better scalability in learning diverse skills that existing RL frameworks have not yet illustrated, affirming its state-of-the-art capability as an answer to question b). We believe that the potential of \method extends significantly beyond the current five skills showcased, suggesting promising avenues for future works.
}

\begin{table*}
\centering
\begin{tabular}{llccccc}
\toprule
\textbf{Goal (Task)} & \textbf{Metric} & \textbf{AMP} & \textbf{AMP w/ H} & \textbf{\dt} & \textbf{TF w/ RHC} & \textbf{\method (Ours)} \\
\midrule
\textbf{$0.3m/s$ Forward}
    & Stability ($\%$) & \textbf{100} & \textbf{100} & 80 & \textbf{100} & \textbf{100} \\
    & $E_v$ ($\%$) & 90.44 $\pm$ 1.87 & 90.63 $\pm$ 4.79 & 75.75 $\pm$ 6.07 & 39.28 $\pm$ 2.34 & \textbf{33.22 $\pm$ 12.48} \\
    % & $\|\omega\|$ (\unit{rad / s}) & 0.12 $\pm$ 0.02 & 0.26 $\pm$ 0.01 & 0.20 $\pm$ 0.05 & 0.46 $\pm$ 0.01 & 0.45 $\pm$ 0.03 \\

\addlinespace
\midrule
\addlinespace

\textbf{$0.5m/s$ Forward}
    & Stability ($\%$) & \textbf{100} & \textbf{100} & \textbf{100} & \textbf{100} & \textbf{100} \\
    & $E_v$ ($\%$) & 50.44 $\pm$ 1.97 & 46.29 $\pm$ 2.55 & 54.35 $\pm$ 2.66 & 37.46 $\pm$ 5.31 & \textbf{12.91 $\pm$ 6.84} \\
    % & $\|\omega\|$ (\unit{rad / s}) & 0.48 $\pm$ 0.01 & 0.42 $\pm$ 0.01 & 0.48 $\pm$ 0.01 & 0.74 $\pm$ 0.04 & 0.62 $\pm$ 0.01 \\

\addlinespace
\midrule
\addlinespace

\textbf{$0.7m/s$ Forward}
    & Stability ($\%$) & 0 & 20 & 0 & 40 & \textbf{100} \\
    & $E_v$ ($\%$) & \cancel{fail 5/5} & 54.96 $\pm$ 0.00 & \cancel{fail 5/5} & 39.36 $\pm$ 5.02 & \textbf{24.80 $\pm$ 8.91} \\
    % & $\|\omega\|$ (\unit{rad / s}) & 0.81 $\pm$ 0.10 & 0.82 $\pm$ 0.14 & 0.88 $\pm$ 0.08 & 1.06 $\pm$ 0.18 & 0.82 $\pm$ 0.01 \\

\addlinespace
\midrule
\addlinespace

\textbf{Turn Left}      
    & Stability ($\%$) & 20 & \textbf{100} & 0. & \textbf{100} & \textbf{100} \\
    & $E_v$ ($\%$) & 20.96 $\pm$ 0.00 & 33.39 $\pm$ 6.96 & \cancel{fail 5/5} & 13.41 $\pm$ 5.02 & \textbf{12.79 $\pm$ 5.64} \\
    % & $\|\omega\|$ (\unit{rad / s}) & 0.65 $\pm$ 0.07 & 0.47 $\pm$ 0.01 & 0.75 $\pm$ 0.09 & 0.65 $\pm$ 0.03 & 0.66 $\pm$ 0.03 \\

\addlinespace
\midrule
\addlinespace
            
\textbf{Turn Right}     
    & Stability ($\%$) & \textbf{100} & \textbf{100} & \textbf{100} & 80 & \textbf{100} \\
    & $E_v$ ($\%$) & 18.61 $\pm$ 2.40 &  33.39 $\pm$ 6.96 & 25.86 $\pm$ 1.47 & 8.69 $\pm$ 5.04 & \textbf{2.22 $\pm$ 1.03} \\
    % & $\|\omega\|$  (\unit{rad / s}) & 0.53 $\pm$ 0.01 & 0.47 $\pm$ 0.52 & 0.53 $\pm$ 0.01 & 0.76 $\pm$ 0.06 & 0.66 $\pm$ 0.02 \\

\addlinespace
\midrule
\addlinespace

\end{tabular}

\caption{Performance Benchmark across different baselines and our \method policy in the real world. Stability (the higher the better) measures the number of trials in which the robot stays stable and does not fall over. $E_v$ (the lower the better) measures the deviation from the desired velocity in percentage. The experiments are conducted with different command settings (Left). Each command is repeated non-stop for five trials, and we report the average and standard deviation of the metrics across five trials.}
\label{tab:performance-metrics}

\end{table*}

\section{\changed{Quantitative Analysis}}\label{sec:quantitative}

In this section, we seek to quantitatively compare \method against various existing multi-skill RL and non-diffusion behavior cloning (BC) baselines. \changed{Since there is no existing RL baseline that learns the five skills in the previous section, we refer to only quadrupedal walking skills, including pacing and trotting, as a case study for the performance of \method.} 

\subsection{Task and Baselines}
This analysis includes walking for four meters under five goals (commands) with different velocities. 
The goals are the following: move forward at three different speeds: 0.3~\unit{m/s}, 0.5~\unit{m/s}, and 0.7~\unit{m/s}, and make a left turn and a right turn at 0.3~\unit{rad/s}. \changed{We record the actual linear velocities via a Kalman filter state estimation~\cite{flayols2017experimental}, and the number of trials where the robot does not fall over through out the trial as the stability metrics. }We repeat each experiment five times and report the mean and standard deviation across five runs. 

\changed{Skill information are often unscalable or unavailable during training and deployment. For a boarder range of applicability, we limit the scope of comparisons within not-skill-conditioned multi-skill RL and non-diffusion BC baselines.} Specifically, the RL baselines include, 

\begin{itemize}
  \item Adversarial Motion Priors (\textbf{AMP})~\cite{escontrela2022adversarial}: An MLP policy trained using AMP with RL (PPO) and style reward of both pacing and trotting reference motions. We \emph{directly} use the \emph{open-sourced} checkpoint from~\cite{escontrela2022adversarial}. \changed{We note that although several skill-conditioned RL policies \cite{wu2023learning, yang2023generalized, li2023versatile} have been introduced since \cite{escontrela2022adversarial}, yielding better sim-to-real results, progress in unconditioned multi-skill policies has been limited.}
  \item AMP with history steps (\textbf{AMP w/ H}): To align with \textbf{\method}, we train an AMP policy with 8 steps of state and action history with the same setup as~\cite{escontrela2022adversarial} and a similar evaluation return in simulation. 
\end{itemize}

Furthermore, we compare \textbf{\method} with non-diffusion BC policies, which can be categorized into autoregressive token prediction~\cite{chen2021decision, huang2023skill} and action sequence prediction as used in \cite{fu2024mobile}. We adopt baselines for each category. 
\begin{itemize}
  \item Transformer with Autoregressive Token Prediction (\textbf{\dt}): A Generative Pretrained Transformer (GPT) \cite{brown2020language} policy similar to a decision transformer~\cite{chen2021decision} without reward conditioning. This only generates one timestep action. 
  
  \item Transformer with Receding Horizon Control (\textbf{TF w/ RHC}): A transformer policy with the same future step action predictions. The model's architecture is identical to our \textbf{\method} model, but it directly predicts future action sequences and the loss is replaced by the reconstruction loss $l = MSE(\pi_\theta(\mathbf{s}_t, \mathbf{g}_t), \mathbf{a}_t)$. 
\end{itemize}
These baselines have the same parameter count of 6.8M and are trained with the same learning rate scheme and number of epochs as \textbf{\method}.

\begin{remark}
    Typically, previous work uses DAgger style algorithms~\cite{ross2011reduction} to better cope with distribution shift, but these methods require access to the expert policy in training and online learning environments. As a more scalable and versatile framework, we limit our focus to learning entirely from offline datasets. 
\end{remark}

\subsection{\method versus AMP (RL)}
\label{sec:diffusion-vs-amp}

We first compare \method with RL-based multi-skill control policy \textbf{AMP}. 
Table~\ref{tab:performance-metrics} shows that \textbf{\method} is the only method among all of the baselines in our benchmark that is able to reliably complete all trials without falling over. 
Specifically, the RL-trained \textbf{AMP} and \textbf{AMP w/ H} baselines struggle with low and high speed commands. 
For 0.3 \unit{m/s} forward command, these two baselines give a velocity tracking performance of more than 90\% slower than the commanded velocity. 
For 0.7 \unit{m/s} forward command, they achieve a stability metric of 0\% and 20\% respectively. 

This shows the prevailing problem of mode-collapse for Generative Adversarial Network (GAN) style networks, such as a multi-skill AMP policy. 
Mode collapse is a significant challenge in GANs, where the generator becomes overfitted to a limited range of outputs that are often similar or identical, rather than offering a broad range of solutions~\cite{nagarajan2017gradient, liu2019spectral, durall2020combating}. 
In the context of AMP, this means the actor network excessively overfits the simulation environment, losing its ability to generalize and adapt to new environments (such as the real world), \changed{but instead reverting to early training behaviors where the discriminator is not yet converged. 
In real-world testing, an example of this is seen in low-speed tasks where AMP oscillates in place with correct frequency but reduced amplitude, similar to its behaviors in early training stage.
In contrast, \method generalizes within the expert demonstrations and does not revert to infeasible actions.} 
Note that in the simulation environment, both \textbf{AMP} and \textbf{AMP w/ H} are able to control the robot to track different velocities without falling over.

Besides better sim-to-real transfer, \textbf{\method} with a diffusion model is able to efficiently learn the multi-modality presented by the different skills for the same locomotion task, and thus able to perform valid and coordinated locomotion skills without mode-collapsing, as an example given by Fig.~\ref{fig:trotting-mode}.
This helps \textbf{\method} achieve both better stability and track completion (velocity tracking) performance compared with AMP-based policy baselines.  

\begin{figure}[t]
  \centering
  \subfloat[]{\includegraphics[width=0.48\linewidth]{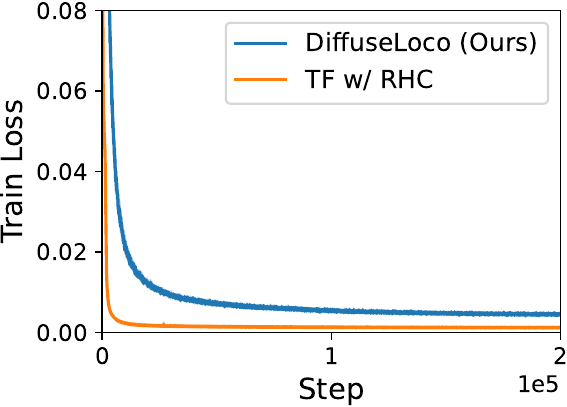}}
  \hfill
  \subfloat[]{\includegraphics[width=0.48\linewidth]{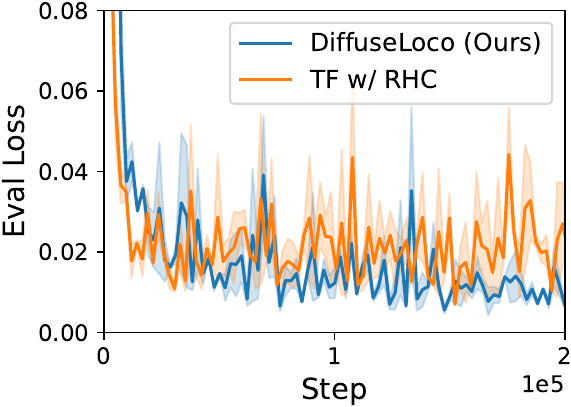}}  
\caption{\label{fig:exp-bcloss} Loss curves for training \textbf{\method} and \textbf{TF w/ RHC} baseline. (a): Training Loss. (b): Evaluation Loss. We report mean and standard deviation across three seeds. We find that even though \textbf{TF w/ RHC} achieves a low reconstruction loss in training, the evaluation loss stays higher than \textbf{\method} and increases at the end. This indicates \textbf{TF w/ RHC} tends to overfit the offline datasets while \textit{DiffuseLoco} is not, with the same number of parameters.}
\end{figure}

\subsection{\method versus Non-Diffusion Behavior Cloning}
\label{sec:diffusion-vs-bc}

We further compare \textbf{\method} with non-diffusion BC methods. 
For locomotion tasks, smooth and temporally consistent actions are a necessity for stability and robustness. 
Looking at Table \ref{tab:performance-metrics}, we find that \textbf{\method} outperforms \textbf{\dt} and \textbf{TF w/ RHC} in both stability and robustness of the locomotion policy. 
Using one-step action output, we find that \textbf{\dt} lacks robustness and fails the 0.7~\unit{m/s} forward and left turn tasks completely. 
This is likely because single-step action prediction makes the policy less aware of future actions, leading to more jittering behavior.

With receding horizon control, \textbf{TF w/ RHC} overcomes most of the jittering problem and can complete most of the tasks. 
However, we note that for more agile motion such as the 0.7 \unit{m/s} forward task, the stability metric drops drastically to merely 40\%. 
This is likely because the reconstruction loss used in \textbf{TF w/ RHC} training tends to overfit the action trajectories in the dataset, resulting in less robust policy in the out-of-distribution scenarios (such as in the real world). 

This is especially evident when looking at training curves for \textbf{TF w/ RHC} versus \textbf{\method} shown in Figure \ref{fig:exp-bcloss}, where \textbf{TF w/ RHC} overfits significantly to the training dataset and the evaluation loss stays high. 
Note that the model architecture is kept identical across \textbf{TF w/ RHC} and \textbf{\method}, so only the loss functions are different. 
In addition, the evaluation loss here is calculated with samples from the same distribution as the training dataset, which means the overfitting problem becomes more pronounced when we switch to real-world experiments, as shown earlier. 

In comparison, our \textbf{\method} shows more stable and smooth motions measured by both stability metrics and magnitudes of the body's angular velocity. 
On average, \textbf{\method} achieves 10.40\% less in magnitude for the body's oscillation over all trials. 
As a result, the smoother locomotion skill helps \textbf{\method} to achieve on average 38.97\% less tracking error compared to \textbf{TF w/ RHC}. 
Based on this observation, we suggest that DDPM-style training is more suitable for imitating locomotion tasks compared to prior Behavior Cloning methods.

\begin{table*}
\centering

\begin{tabular}{llcccccc}
\toprule
\textbf{Goal (Task)} & \textbf{Metric} & \textbf{DL w/o RHC} & \textbf{DL w/o Rand} & \textbf{\ddim} & \textbf{\ddimn} & \textbf{\changed{U-Net}} & \textbf{\method (Ours)} \\
\midrule
\textbf{$0.3m/s$ Forward}
    & Stability ($\%$) & \textbf{100} & \textbf{100} & \textbf{100} & \textbf{100} & \changed{\textbf{100}} & \textbf{100} \\
    & $E_v$ ($\%$) & 75.09 $\pm$ 18.98 & 50.45 $\pm$ 2.70 & 56.89 $\pm$ 2.43 & 47.09 $\pm$ 2.40 & \changed{81.31 $\pm$ 1.90} & \textbf{33.22 $\pm$ 12.48}\\
    % & $\|\omega\|$ (\unit{rad / s}) & 0.26 $\pm$ 0.18 & 0.46 $\pm$ 0.00 & 0.20 $\pm$ 0.05 & 0.44 $\pm$ 0.02 & 0.43 $\pm$ 0.01 \\

\addlinespace
\midrule
\addlinespace

\textbf{$0.5m/s$ Forward}
    & Stability ($\%$) & \textbf{100} & 80 & 80 & \textbf{100} & \changed{\textbf{100}} & \textbf{100} \\
    & $E_v$ ($\%$) & 64.49 $\pm$ 1.87 & 41.07 $\pm$ 6.12 & 41.00 $\pm$ 3.18 & 37.92 $\pm$ 1.59 & \changed{74.52 $\pm$ 2.83} & \textbf{12.91 $\pm$ 6.84} \\
    % & $\|\omega\|$ (\unit{rad / s}) & 0.57 $\pm$ 0.02 & 0.63 $\pm$ 0.03 & 0.48 $\pm$ 0.01 & 0.68 $\pm$ 0.04 & 0.61 $\pm$ 0.02 \\

\addlinespace
\midrule
\addlinespace

\textbf{$0.7m/s$ Forward}
    & Stability ($\%$) & 0 & 40 & 80 & 80 & \changed{\textbf{100}} & \textbf{100}\\
    & $E_v$ ($\%$) & \cancel{fail 5/5} & 44.30 $\pm$ 4.21 & 47.71 $\pm$ 6.63 & 42.58 $\pm$ 2.08 & \changed{71.71 $\pm$ 2.93} & \textbf{24.80 $\pm$ 8.91}  \\
    % & $\|\omega\|$ (\unit{rad / s}) & 0.83 $\pm$ 0.04 & 0.81 $\pm$ 0.03 & 0.84 $\pm$ 0.06 & 0.83 $\pm$ 0.02 & 0.84 $\pm$ 0.09 \\

\addlinespace
\midrule
\addlinespace

\textbf{Turn Left}      
    & Stability ($\%$) & \textbf{100} & \textbf{100} & \textbf{100} & \textbf{100} & \changed{20} & \textbf{100}\\
    & $E_v$ ($\%$) & 20.96 $\pm$ 18.22 & \textbf{10.17 $\pm$ 5.86}  & 22.22 $\pm$ 4.29 & 13.27 $\pm$ 2.63 & \changed{18.93 $\pm$ 23.28} & 12.79 $\pm$ 5.64 \\
    % & $\|\omega\|$ (\unit{rad / s}) & 0.65 $\pm$ 0.07 & 0.69 $\pm$ 0.01 & 0.75 $\pm$ 0.09 & 0.67 $\pm$ 0.03 & 0.63 $\pm$ 0.02 \\

\addlinespace
\midrule
\addlinespace
            
\textbf{Turn Right}     
    & Stability ($\%$) & \textbf{100} & \textbf{100} & \textbf{100} & \textbf{100} & \changed{\textbf{100}} & \textbf{100} \\
    & $E_v$ ($\%$) & 18.61 $\pm$ 2.40 &  8.18 $\pm$ 3.94 & 6.47 $\pm$ 2.49 & 7.42 $\pm$ 2.90 &  \changed{89.63$\pm$ 3.36} & \textbf{2.22 $\pm$ 1.03} \\
    % & $\|\omega\|$  (\unit{rad / s}) & 0.53 $\pm$ 0.01 & 0.73 $\pm$ 0.02 & 0.53 $\pm$ 0.01 & 0.69 $\pm$ 0.03 & 0.67 $\pm$ 0.01 \\

\addlinespace
\midrule
\addlinespace

\end{tabular}

\caption{Performance Ablation Study across different ablations and \method policy in real-world experiments. Stability (the higher the better) measures the number of trials in which the robot stays stable and does not fall over. $E_v$ (the lower the better) measures the deviation from the desired velocity in percentage. The experiments are conducted with different command settings (Left). Each command is repeated non-stop for five trials, and we report the average and standard deviation of the metrics across five trials.}
\label{tab:ablation-study}

\end{table*}

\section{Ablation Study on Design Choices}
\label{sec:ablation-study}
In this section, we further evaluate the design choices used to build \method policy in simulation and the real world by extensive ablation studies. We use the same experiment setups as the previous section. \changed{For brevity of the main content, further ablation studies can be found in Appendix \ref{appx:more-ablation}}.

\subsection{Ablation Components}
To validate our design choices, we ablate \textbf{\method} with the following critical components and compare them to our real-world benchmark. 

\begin{itemize}
  \item Without Receding Horizon Control (\textbf{DL w/o RHC}): Replace RHC with one-step prediction in an autoregressive manner and keep the diffusion model. 

  \item Without Domain Randomization (\textbf{DL w/o Rand}): Trained on a dataset generated without domain randomization, except for the ground friction coefficient.

  \item DDIM Inference: 
   We develop two DDIM baselines to investigate how training and inference steps affect performance in locomotion control. 
  \begin{itemize}
        \item 100 Training + 10 Inference (\textbf{\ddim})
        \item 10 Training + 5 Inference (\textbf{\ddimn})
  \end{itemize}
  Compared with our \textbf{\method}, \textbf{\ddim} has the same inference steps, and \textbf{\ddimn} has the same training steps. 

  \item \changed{U-Net as the backbone (\textbf{U-Net}): Replace the Transformer with a U-Net as the backbone, adjusted to the same parameter count.} 
\end{itemize}

\subsection{Single-step output versus RHC}\label{exp:rhc}
To isolate the effects of RHC, we test a variant of \method without RHC (\textbf{DL w/o RHC}), finding that it struggles with faster speed goal and exhibits significant jittering behaviors, as detailed in Table~\ref{tab:ablation-study}. This suggests that single-step token-prediction models like GPT are less suitable for legged locomotion control than diffusion models, which predict sequences of future actions.

\subsection{\changed{Sampling Techniques}}
\label{exp:ddpmvsddim}
As discussed earlier, popular diffusion-based frameworks like DDIM often reduce sample iterations for inference acceleration, trading off output quality for speed, often with ten times fewer iterations~\cite{song2022denoising}. 
While this approach suits tasks like image generation, which tolerate some variance, it underperforms in quadrupedal locomotion control. As shown in Table \ref{tab:ablation-study}, both \textbf{\ddim} and \textbf{\ddimn} exhibit worse stability and higher velocity tracking errors. Noticeably, the 100 training steps and 10 inference steps variant demonstrates limping behavior and fails two trials. Tracking errors for both variants increase by 50.69\% and 42.04\%, respectively, compared to \textbf{\method}.

Thus, we believe that noisier control signals from the DDIM pipeline likely disrupt the control of inherently unstable floating-based dynamic systems, like legged robots. \changed{An interesting future work direction could be on control-specific sampling techniques to accelerate diffusion models without compromising stability and performance.}

\subsection{\changed{Model Architecture Effects}}
\changed{
In addition, we compare against another commonly used architecture in diffusion models, a CNN-based U-Net as the backbone of \method. Qualitatively, the \textbf{U-Net} policy is shaky and inconsistent, and quantitatively, it has one of the highest errors in velocity tracking, with worsened stability due to its shaky actions. We reckon that this is because CNNs are not the best fit for temporal data and also lack separate attention weights for goal conditioning. This is consistent with prior work~\cite{chi2023diffusion} that finds \textbf{U-Net} underperforms Transformer, especially in high action-rate dynamical systems.
}

\subsection{Dataset Effects}
\label{exp:norand}
Lastly, we explore how dataset characteristics influence the robustness and performance of \textbf{\method} in real-world scenarios. Consistent with previous findings that diversity in training data, such as noise insertion, mitigates compounding error~\cite{laskey2017dart}, we demonstrate that increasing the variety of dynamics parameters in simulation environments 
 where we collect data also enhances robustness. As in Table \ref{tab:ablation-study}, training \method on a dataset with dynamics randomization leads to a 44.26\% increase in both robustness and stability compared to \textbf{DL w/o Rand} baseline. Specifically, in the challenging 0.7~\unit{m/s} forward task, \textbf{DL w/o Rand} falls in 3 out of 5 trials. This ablation study points to the potential of altering the dataset, by adding either more diversity and potentially real-world data or more fault-recovery behaviors, to further enhance the robustness of \method.
\label{sec:Results}

\section{Discussion and Future Work}
\label{sec:discussion}

We have presented \method, a scalable learning framework to learn diverse agile legged locomotion skills from multi-modal \emph{offline} datasets and can robustly transfer to real-world robots in real-time.
Leveraging diffusion models to capture the multi-modality in the offline dataset, \method learns a state-of-the-art controller that combines bipedal walking and quadrupedal locomotion skills within one policy and transitions freely among the skills.  

\subsection{\changed{A Scalable Approach for Locomotion Skills}}
\changed{
In this work, we focus on the scalability of learning legged locomotion control. Inspired by successes of large-scale learning in other robotics tasks, we leverage the most scalable approach: learning from offline datasets, with special attention to the versatility of the data sources and the multi-modality of different skills in a large-scale dataset. With an expressive diffusion model as the backbone, we are able to absorb demonstrations learned with various existing RL algorithms with potentially different observation and action spaces, and effectively execute skills, such as trotting and pacing, that represent different modalities given identical commands.} With the five diverse skills presented in this work as a testimony, we show the scalability of \method towards a generalist policy for locomotion control tasks.

\subsection{Benefits over Other Multi-skill Policies}
As we show in our thorough real-world benchmark (Sec. \ref{sec:quantitative}), \method demonstrates smoother actions and improved stability and velocity tracking in real-world conditions compared to non-diffusion Behavior Cloning baselines that are commonly used in prior works. When not conditioned on explicit skill labels, \method shows better sim-to-real transfer performance for multi-skill locomotion compared to AMP policies, which often face a significant sim-to-real gap due to mode-collapsing in generator-discriminator methods. \method avoids these issues, providing stable, coherent, and effective control with smooth skill-switching and stable execution under consistent commands.

\subsection{Large-scale Offline Dataset for Locomotion}

In the experiments, \method demonstrates scalability and robustness by utilizing diverse offline datasets, as discussed in Sec. \ref{exp:norand}. 
Unlike online learning which mostly depends on simulations, it offers a practical inspiration for scalable real-world data collection and learning. \changed{Similar to \cite{feng2023genloco}, we also hypothesize that \method could adapt to datasets containing different robot morphologies, thus allowing for broader deployment and better generalization. Furthermore, as a popular direction in manipulation fields, integrating vision and language instructions into the goal-conditioning dataset could further enhance \method's versatility and applicability in future works. 
}

\section{Acknowledgement}
This work was supported in part by NSF 2303735 for POSE, in part byNSF 2238346 for CAREER, in part by The AI Institute and in part byInnoHK of the Government of the Hong Kong Special AdministrativeRegion via the Hong Kong Centre for Logistics Robotics.
\label{sec:Conclusion}

% \section*{Acknowledgments}

%% Use plainnat to work nicely with natbib. 

\bibliographystyle{plainnat}
\bibliography{references}

\clearpage
\begin{appendices}

\onecolumn
\section{\changed{More Results on Skill Transitioning}}
\label{appx:transitioning}

% In this section we present more results on a series of video clips to showcase skill transitioning of \method.

\begin{figure}[!ht]
\centering
\includegraphics[width=\linewidth]{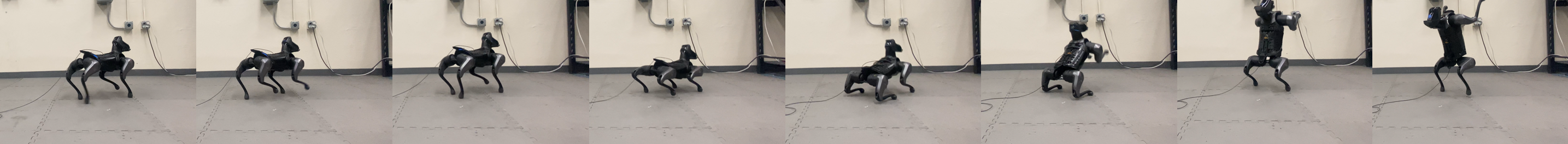}
\setlength{\abovecaptionskip}{-8pt}
\setlength{\belowcaptionskip}{-12pt}
\caption{\label{fig:pace_to_stand}\changed{Skill Transitioning: Pace to Stand}}
\end{figure}

\begin{figure}[!ht]
\centering
\includegraphics[width=\linewidth]{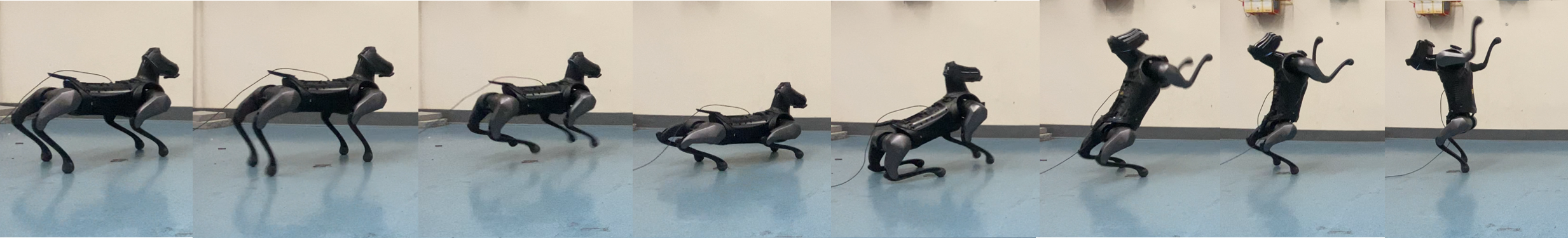}
\setlength{\abovecaptionskip}{-8pt}
\setlength{\belowcaptionskip}{-12pt}
\caption{\label{fig:hop_to_stand}\changed{Skill Transitioning: Hop to Stand on Bare Floor}}
\end{figure}

\begin{figure}[!ht]
\centering
\includegraphics[width=\linewidth]{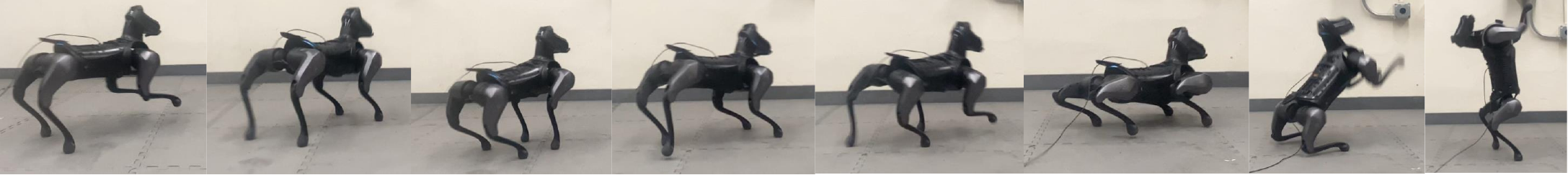}
\setlength{\abovecaptionskip}{-8pt}
\setlength{\belowcaptionskip}{-12pt}
\caption{\label{fig:bounce_to_stand}\changed{Skill Transitioning: Bounce to Stand with Emergent Intermediate Pacing Skill}}
\end{figure}

\begin{figure}[!ht]
\centering
\includegraphics[width=\linewidth]{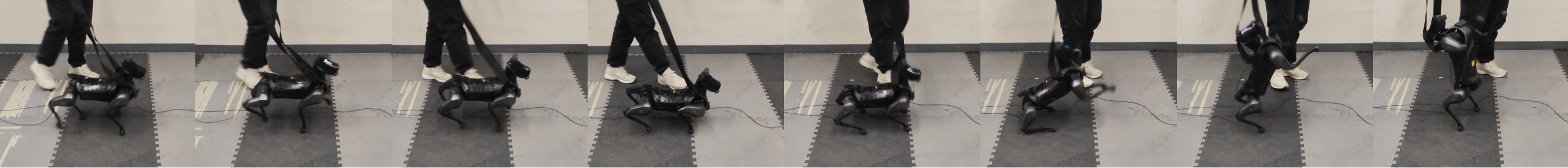}
\setlength{\abovecaptionskip}{-8pt}
\setlength{\belowcaptionskip}{-12pt}
\caption{\label{fig:trot_to_stand}\changed{Skill Transitioning: Trot to Stand}}
\end{figure}

\begin{figure}[!ht]
\centering
\includegraphics[width=\linewidth]{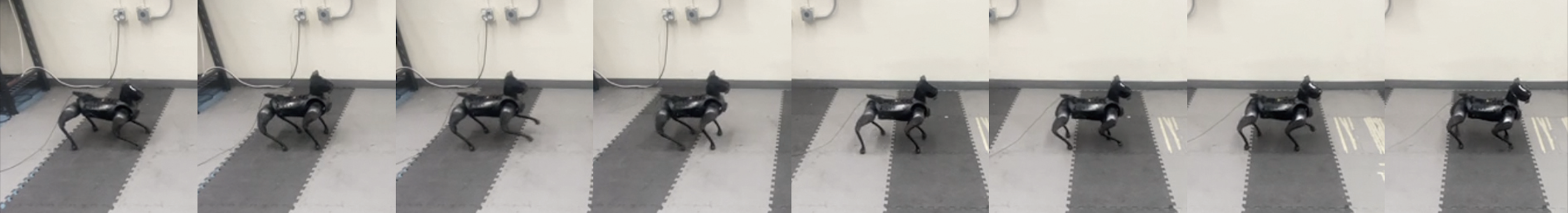}
\setlength{\abovecaptionskip}{-8pt}
\setlength{\belowcaptionskip}{-12pt}
\caption{\label{fig:bounce_to_pace}\changed{Skill Transitioning: Bounce to Pace}}
\end{figure}

\begin{figure}[!ht]
\centering
\includegraphics[width=\linewidth]{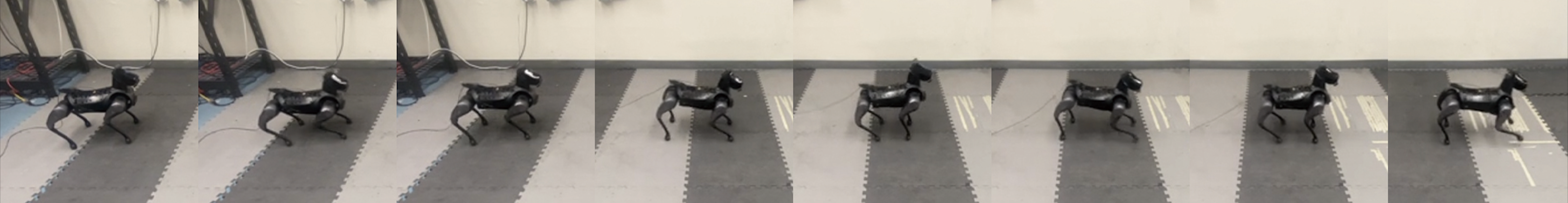}
\setlength{\abovecaptionskip}{-8pt}
\setlength{\belowcaptionskip}{-12pt}
\caption{\label{fig:hop_to_bounce}\changed{Skill Transitioning: Hop to Bounce}}
\end{figure}

\begin{figure}[!ht]
\centering
\includegraphics[width=\linewidth]{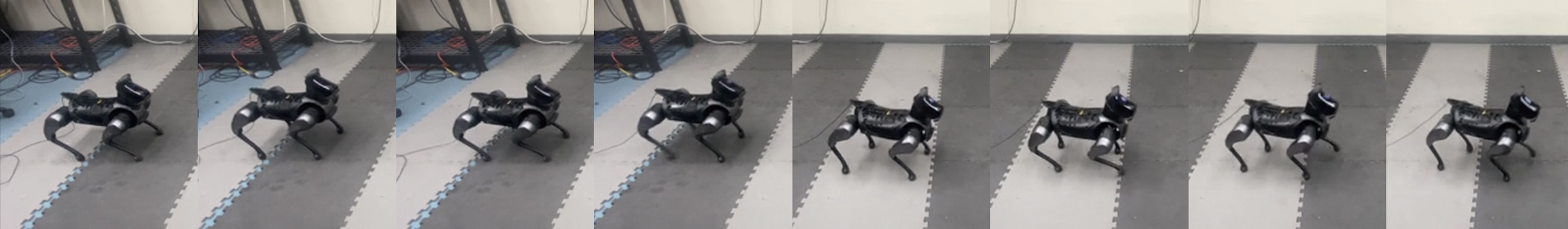}
\setlength{\abovecaptionskip}{-8pt}
\setlength{\belowcaptionskip}{-12pt}
\caption{\label{fig:hop_to_pace}\changed{Skill Transitioning: Hop to Pace}}
\end{figure}

\twocolumn

\section{Model Architecture and Hyperparameters}
\label{appx:architecture}

Here, we explain the details of the diffusion model's architecture, dataflow, and transformer backboone. 

\subsection{Receding Horizon Control}
\label{appx:rhc}
\changed{Learning sequences of actions instead of single-step action in training helps improving temporal consistency of the policy.} 
However, in dynamic systems such as legged robots, error accumulates significantly after a short horizon of planned steps, and the predicted actions further ahead may no longer be useful for control.  
Therefore, we adopt the Receding Horizon Control (RHC) manner, where \method policy generates $n$ steps of actions, but only executes the \emph{very first step} of actions.
This is in contrast to previous work which infers a sequence of actions at a lower frequency and using interpolation to get a high-frequency action \cite{kapelyukh2023dall, li2024crossway}.
Such a setup allows us to replan rapidly with fast-changing states of the robot while keeping future steps in account.
As we evaluated in Sec.~\ref{exp:rhc}, using RHC is critical in improving smoothness and consistency of a legged locomotion control policy.

\subsection{Architecture}
The \method policy leverages an encoder-decoder transformer DDPM. First, the past robot's past I/O trajectory $(\mathbf{s}_{t-h-1:t-1}, \mathbf{a}_{t-h-2:t-2})$ and given goal sequence $\mathbf{g}_{t-h-1:t-1}$ are transformed into separate I/O embedding and goal embedding by two 2-layer MLP encoders, respectively. 
Then, we sample noise $\epsilon(k)$ for diffusion time step $k$ with the DDPM scheduler and add to the ground truth action $\mathbf{a}$ from the offline dataset to produce a noisy action $\mathbf{a}^k_{t:t+n} = \mathbf{a}_{t:t+n} + \epsilon_k$. 
The noisy action $\mathbf{a}^k_{t:t+n}$ is then passed through an MLP layer into action embedding.  
The noisy action tokens are then passed through 6 Transformer decoder layers, each of which is composed of an 8-head cross-attention layer. 
Each layer computes the attention weights for the noisy action tokens querying all the state embedding, goal embedding, and the timestep embedding reflecting the current diffusion timestep $k$. 
We apply causal attention masks to each of the state embeddings and goal embeddings separately. The predicted noise $\epsilon_\theta(\mathbf{a}_{t-h-2:t+n}, \mathbf{s}_{t-h-1:t-1}, \mathbf{g}_{t-h-1:t-1}, k)$ is then computed by each corresponding output token of the decoder stack. We then supervise the output to predict the added noise with Eqn.~\ref{eqn:ddpm-loss-goal-delay} to find optimal parameters $\theta$ of the denoising model $\epsilon_\theta$.

\subsection{\changed{Hyperparameters}}
\changed{The hyperparameters are summarized in Table \ref{tab:transposed_hyperparameters}, }

\begin{table}[ht]
\centering
\caption{\changed{Hyperparameters for \method in the Experiments}}
\label{tab:transposed_hyperparameters}
\begin{tabular}{lccc}
\hline
& \begin{tabular}[c]{@{}c@{}}Five-Skill\\ (Sec. \ref{sec:multi-skill})\end{tabular} & \begin{tabular}[c]{@{}c@{}}Walk\\ (Sec. \ref{sec:quantitative})\end{tabular} & \begin{tabular}[c]{@{}c@{}}Cassie\\ (Sec. \ref{sec:biped})\end{tabular} \\ \hline
History Length & 8 & 8 & 16 \\
Predict Length & 4 & 4 & 4 \\
Token Dim & 256 & 128 & 256 \\
Attn Drop-out & 0.3 & 0.3 & 0.3 \\
Learning Rate & 1e-4 & 1e-4 & 1e-4 \\
Weight Decay & 1e-3 & 1e-3 & 1e-3 \\
Epochs & 100 & 100 & 100 \\
\hline
\end{tabular}
\end{table}

\section{More Ablation Studies on Design Choices}
\label{appx:more-ablation}

\begin{figure}[ht]
\centering
\includegraphics[width=0.8\linewidth]{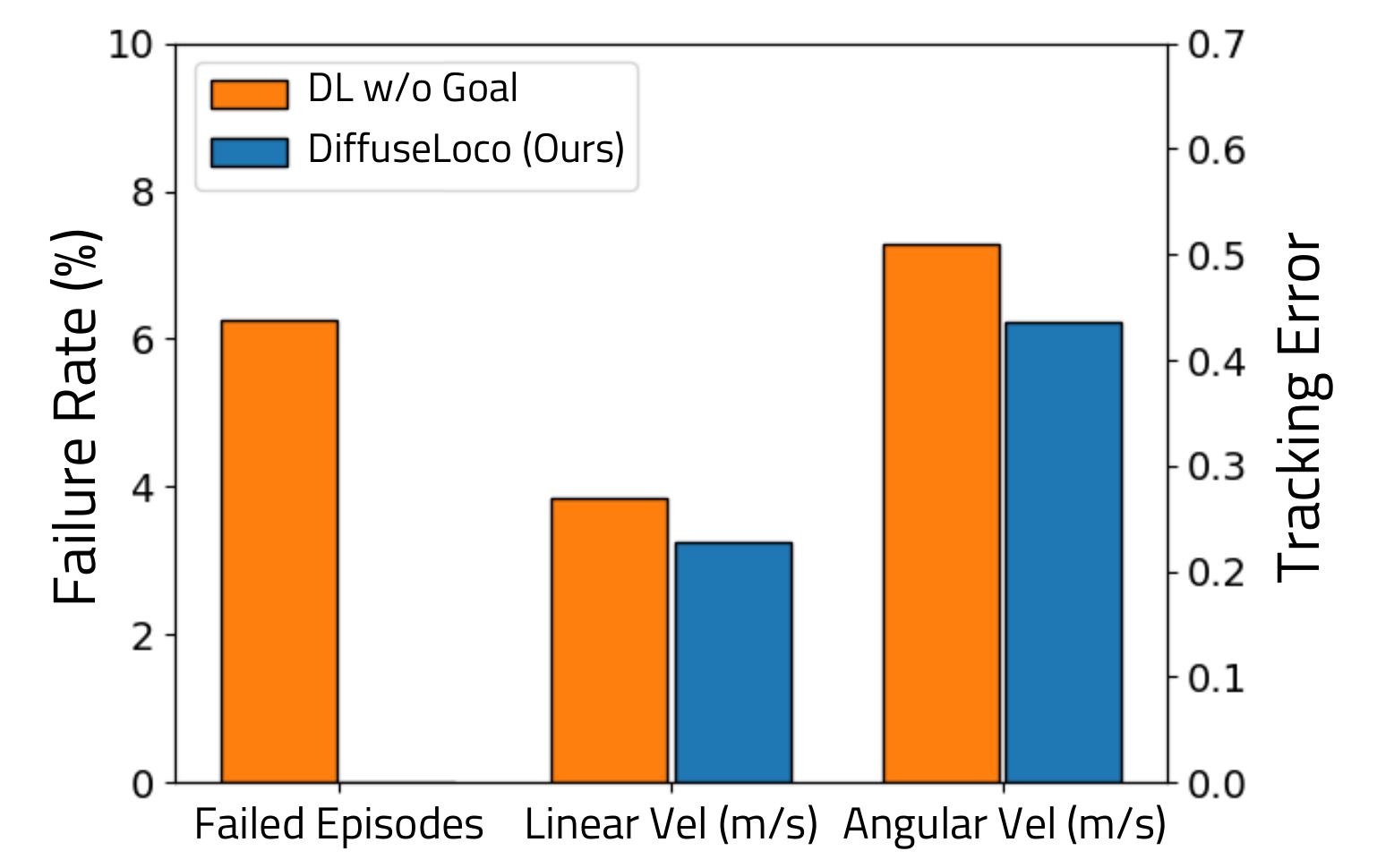}
\caption{\label{fig:nogoal-metrics}Comparison of failure rates and tracking errors between \textbf{DL w/o Goal} and \textbf{DiffuseLoco} (ours) in simulation. The left y-axis is the metric for Failed Episodes. The right y-axis indicates the tracking error for linear velocity and angular velocity.}
\end{figure}

\subsection{Use of Goal-conditioning}
\label{exp:goal-cond}

Here, we evaluate the impact of goal-conditioning, hypothesizing enhancements tracking performance and stability. 
In the $\textbf{DL w/o Goal}$ baseline, we do not add the goal-conditioning encoder. Instead, the goal is concatenated with the robot's I/O. 
Considering the noisy base velocity estimations on real robots, we utilize simulation environments with extensive dynamics randomization to perform a large number of trials and get more systematic results. 
As shown in Fig.\ref{fig:nogoal-metrics}, \method achieves a 15.4\% reduction in linear velocity tracking error and a 14.5\% reduction in angular velocity error compared to the \textbf{DL w/o Goal} baseline. Moreover, over 64 trials with identical commands, \textbf{DL w/o Goal} falls over four times, or 6.25\% of all trials, whereas \method experiences no failures. This pattern persists in real-world testing, where \textbf{DL w/o Goal} fails one trial in a 0.7\unit{m/s} forward test.

These results underscore the importance of goal-conditioning with distinct attention weights for dynamic system control, revealing that the robot's I/O history and goals, governed by physics and arbitrary objectives respectively, should not be merged into one embedding space. 

\section{Details in Offline Locomotion Dataset}
\label{appx:dataset}

Here, we will introduce the details in creating the offline locomotion dataset used in this work. 
We will explain the state, goal, and action spaces, followed by a brief introduction to the source policies and dynamics randomization used to diversify the data. We collect a total of 4 million data of state-action-goal pairs in the offline dateset for the quadrupedal robot tasks, and 10 million transitions for the bipedal robot tasks.

\subsection{State Space}

The state space is the robot's proprioceptive feedback. 
In the quadrupedal locomotion control case, this consists of the measured motor positions $\mathbf{q}_m$, measured motor velocities $\dot{\mathbf{q}}_m$, base orientation $\mathbf{q}_{\psi,\theta,\phi}$, and base angular velocities $\dot{\mathbf{q}}_{\psi,\theta,\phi}$. 
Note that we exclude quantities from the estimation of base velocity ($\dot{\mathbf{q}}_{x,y}$) to prevent additional estimation errors. 

\subsection{Goal Space}

The goal of the locomotion task is the commands given to the policy. 
For quadrupedal robots, the command includes desired sagittal velocity $q^d_x$ in the range of 0~\unit{m/s} to 1~\unit{m/s}, desired base height from 0.2~\unit{m} to 0.6~\unit{m}, and desired turning velocity $q^d_\psi$ from -1~\unit{rad/s} to 1~\unit{rad/s}. 

\subsection{Action Space}

The action space is the robot's joint-level commands. In this work, we use the desired motor position $\mathbf{q}^d_m$ as the action. 
This is then used by joint-level PD controllers to compute motor torques $\tau$ at a higher frequency. 

\subsection{Source Policy}

We obtain source policies using three different RL methods. We leverage proximal policy optimization (PPO)~\cite{schulman2017proximal} to optimize each of the source policies, and we train the policies in simulation (Isaac Gym \cite{rudin2022learning}).
We evenly distribute the data generated from each of the source skill-specific policies.

\subsubsection{\changed{Adversarial Motion Prior (AMP)}}
\changed{
For skills trained with AMP, we provide a reference motion retargeted from motion capture data of a dog~\cite{Peng2020Learning}, and incorporate an GAN-style discriminator to encourage the robot to imitate the reference motion without extended reward engineering. 
Then, the reward of this method is formulated as motion imitating term (provided by the discriminator \cite{peng2021amp}) and task term (\emph{e.g.}, tracking error, etc). }

\subsubsection{\changed{Central Pattern Generator Guidance (CPG)}}
\changed{
For skills trained with CPG-guidance, we follow the formulation in \cite{shao2021learning} and provide nominal reference motions of strictly periodic motions generated by a Central Pattern Generator with phase signals. Specifically, for hopping skill, the phase selections for all legs are 0, and for bouncing skill, the phase selections are 0 for the front legs, and $\pi$ for the hinder legs.  
The reward is composed of task term (\emph{e.g.}, velocity tracking error, etc), motion tracking term (\emph{e.g.}, reference motion tracking error, etc), and smoothing terms (\emph{e.g.}, action rate, etc).}  

\subsubsection{\changed{Symmetry Augmented RL}}
\changed{
We train the bipedal locomotion skill for quadrupedal robots following a symmetry-augmented RL policy~\cite{su2024leveraging} to achieve a symmetric gait pattern that is crucial for sim-to-real transfer. 
Specifically, the data collection process is augmented by the addition of symmetric states and actions. 
The reward includes task term (\emph{e.g.}, velocity tracking error, etc), gait pattern term (\emph{e.g.}, feet clearance height, etc), and smoothing terms (\emph{e.g.}, action rate, etc). }

\subsubsection{\changed{Other Methods}}
\changed{
Although the source policies used to collect the dataset in this work are all RL-based policies, our framework is general and can include the data generated from model-based optimal controllers (such as from~\cite{liao2023walking}) and others. The requirement is to align the state and action spaces among different source policies, and the frequency of the policy should be kept the same. }

\subsection{Dynamics Randomization}

In order to diversify the training dataset for \method, we also include the same amount of dynamics randomization~\cite{peng2018sim} during the training of the source policies and the data generation using these policies. 
Specifically, in each episode in simulation, the dynamics parameters are randomized. 
These include the motor's PD gains, the mass of the robot's base \changed{(up to the weight of the onboard compute)}, ground frictions, and random changes in base velocity. 
The randomization ranges are adapted from the source policy's original methods.

\section{Real-Time Inference Acceleration}
\label{appx:acceleration}

Although the diffusion model targets real-time use, it cannot meet the real-time targets without further tuning. Compared to previous works using Transformer for locomotion control with 2M parameters~\cite{radosavovic2023learning}, our model is 3 times larger in parameter count(6.8M parameters) and needs to be forwarded 10 times in each inference. Thus, an additional effort is needed to accelerate the diffusion on the edge computing device on the robot, such as the setup shown in Fig.~\ref{fig:on-board-compute}. 
In this section, we explore several methods to accelerate the inference process of the diffusion model to enable it to run real-time onboard. 

\subsection{Acceleration Framework}

\begin{figure}[t]
\centering
\includegraphics[width=\linewidth]{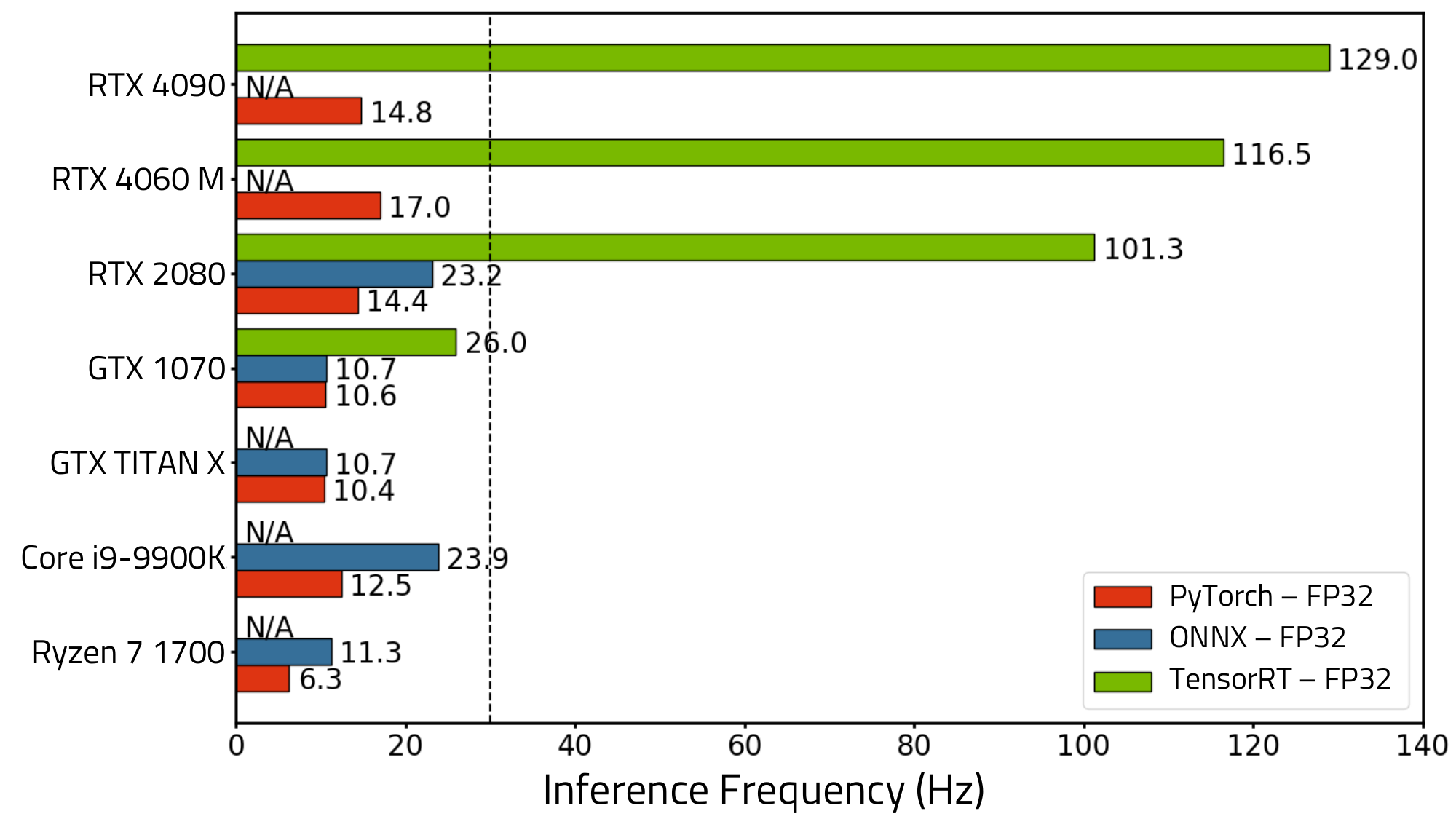}
\caption{\label{fig:hardware-benchmark}Benchmark of running our \method policy (6.8M parameters) on different hardware platforms. The dashed line remarks the 30~\unit{Hz} minimum frequency required to control the robot in real time. We utilize TensorRT to optimize the computation graph and achieve approximately 7 times speedup of inference computation time compared to the naive PyTorch implementation. The N/A entries are due to the lack of software compatibility on the corresponding platform.}
\end{figure}

Our \method policy has a parameter count of 6.8 million parameters, which exceeds most modern mobile processors' cache capacity. 
Furthermore, hardware on a typical consumer-grade central processing unit (CPU) is not optimized for the operators used in transformer networks. 
The graphics processing unit (GPU) is more suitable for computing the high-dimension matrix and vector operations. 
To ensure the portability of the setup, we use an accessible NVIDIA Mobile GPU as the deployment platform.
For real-time deployment, an acceleration pipeline is built in the \method framework to convert and optimize our model towards the target compute platforms. 
The operators of the model are first extracted with ONNX~\cite{onnx}. 
Then, TensorRT is used to refine the execution graph and compile the resulting execution pipeline onto the target GPU. 
Through domain-specific architecture optimizations, the operations and memory access patterns are optimized to utilize the full capability of the GPU. 
With this approach, the speed for each denoising iteration is increased by about 7X compared to the native implementation in PyTorch, and the maximum inference (with 10 denoising iterations) frequency is increased from 17.0~\unit{Hz} to 116.5~\unit{Hz}. To showcase the effect of this acceleration approach, we conducted a benchmark on the inference frequency of the policy running on multiple hardware platforms we have access to, shown in Fig.~\ref{fig:hardware-benchmark}.

\subsection{Edge Compute for \method Policy on Robots}

\begin{figure}[t]
\centering
\includegraphics[width=0.6\linewidth]{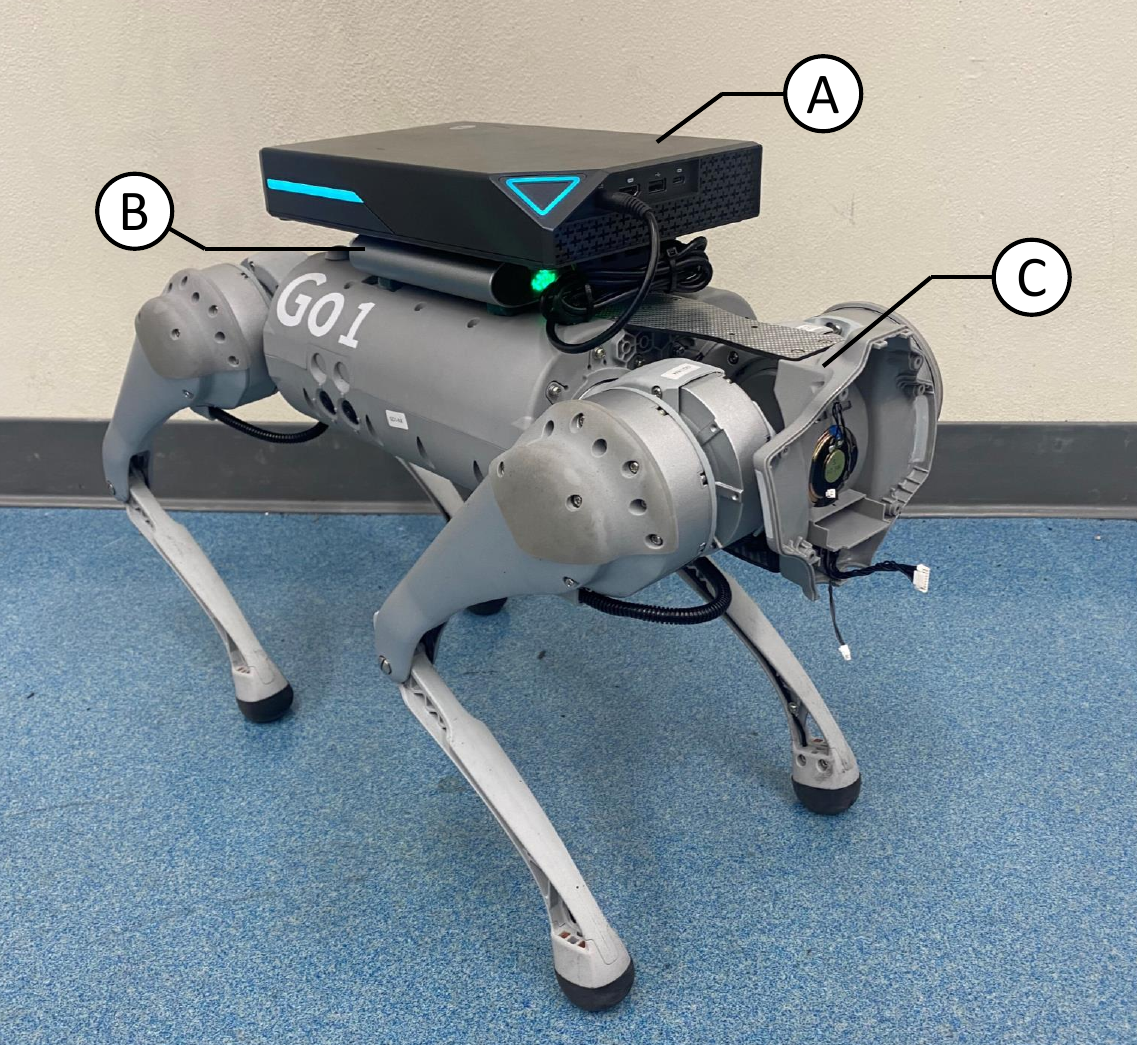}
\caption{\label{fig:on-board-compute} Onboard compute experiment setup. A: Mini computer with Intel Core i7-13700H and NVIDIA GeForce RTX 4060 Mobile. B: Battery bank. C: Go1 quadrupedal robot. This setup is below the robot's adaptive load capacity and the robot can walk with our policy steadily. With our acceleration framework, we are able to achieve onboard and real-time deployment of our diffusion model.}
\end{figure}

With the help of the acceleration framework, the compute platform can be deployed onboard a Go1 quadruped robot. A mini-computer equipped with Intel Core i7-13700H and NVIDIA GeForce RTX 4060 Mobile is attached to the top of the robot, as showcased in Fig.~\ref{fig:on-board-compute}. This computer runs \method policy and is powered by a dedicated battery bank, separated from the robot's internal battery. This arrangement is capable of running the policy for up to 90 minutes. 
The mini-computer connects to the robot via an Ethernet cable to send action for the joint-level PD controls on the robot's computer. We note that all the experiments we present are completed on this edge computing device. 

\begin{figure}[ht]
\centering
\includegraphics[width=\linewidth]{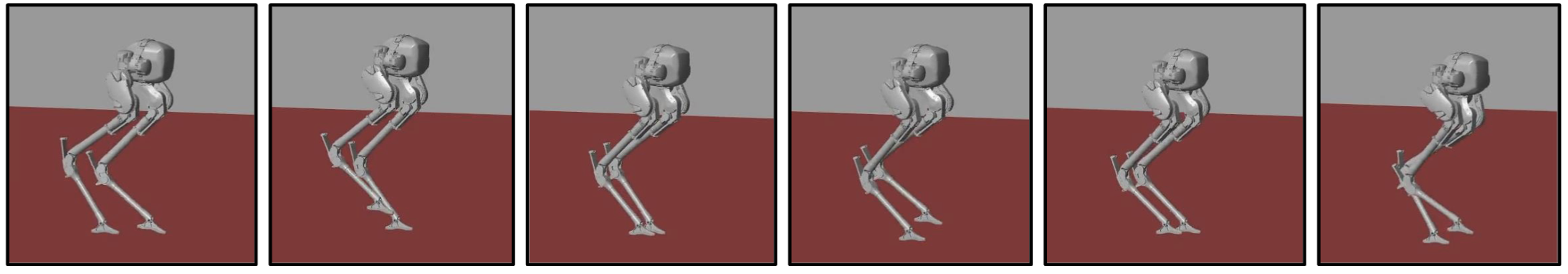}
\includegraphics[width=\linewidth]{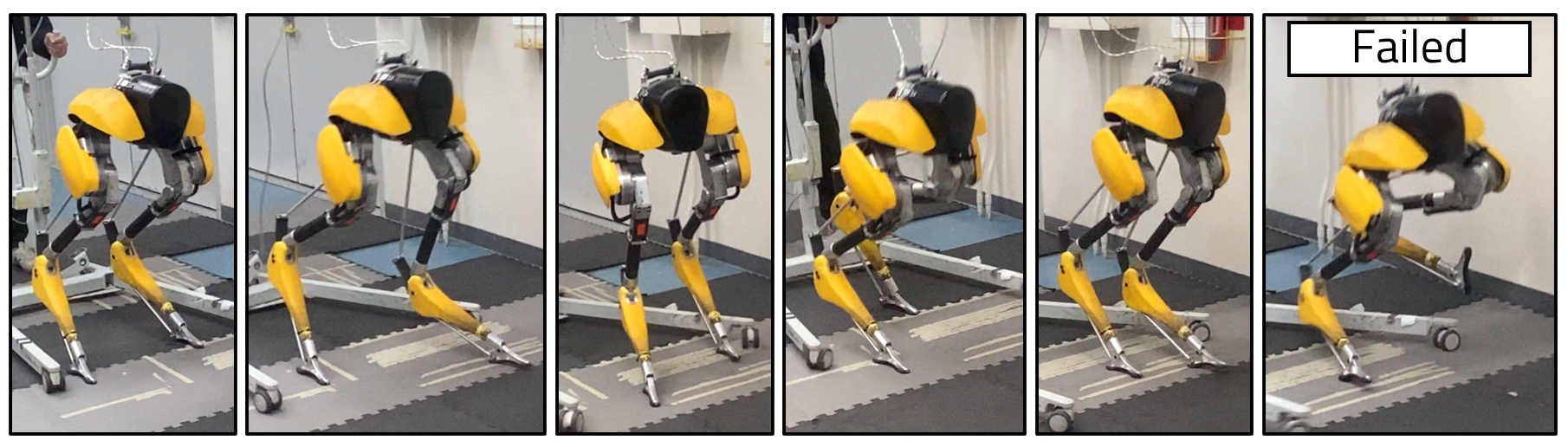}
\caption{\label{fig:real-world-cassie} Snapshots of \changed{(top) deploying \method policy on a high-fidelity Simulink simulation and} (bottom) the Cassie hardware. Although working in sim-to-sim transfer, the \method policy encounters difficulty in zero-shot transferring on the real Cassie hardware due to a much larger sim-to-real gap. Extending \method to a higher dimensional more complex dynamic system in the real world as an important future work.}
\end{figure}

\section{Limitations}
\label{appx:limitations}

In the experiments, our \method policy demonstrates good robustness against various ground conditions and small variations in terrain. However, against skill-specific RL policies, the robustness of our multi-skill \method policy is insufficient. 
For instance, a quadrupedal robot controlled by the \method policy struggles with recovery from large external perturbations, while skill-specific RL policies handle effectively. 
This limitation becomes more evident where our policy fails to overcome the significant sim-to-real gap on Cassie.
Although the diffusion policy for Cassie worked in the high-fidelity simulation (Simulink), the policy only sustained a few steps in hardware experiments before Cassie lost balance and fell down as shown in Fig. ~\ref{fig:real-world-cassie}.

\end{appendices}

\end{document}